\documentclass[final,5p,times,twocolumn]{elsarticle}

\usepackage{setspace}
\usepackage{pifont}

\usepackage{algorithm}
\usepackage{algpseudocode}
\usepackage{graphicx}

\usepackage{subfigure}
\usepackage{amsmath}

\usepackage{amssymb}
\usepackage{xcolor}
\usepackage{tabularx} 
\usepackage{multirow}
\usepackage{tabu}
\usepackage{pbox}
\usepackage{amssymb}
\usepackage{amsfonts}
\usepackage{bbm}
\usepackage{graphicx}
{\begin{list}               
    {$\bullet$ \hfill}{
        \setlength{\leftmargin}{\parindent}
        \setlength{\parsep}{0.04\baselineskip}
        \setlength{\itemsep}{0.5\parsep}
        \setlength{\labelwidth}{\leftmargin}
        \setlength{\labelsep}{0em}}
    }
{\end{list}}

\providecommand{\cref}[1]{Chapter~\ref{#1}}

\providecommand{\fref}[1]{Figure~\ref{#1}}
\providecommand{\tref}[1]{Table~\ref{#1}}

\providecommand{\norm}[1]{\lVert#1\rVert}

\providecommand{\mat}[1]{\ensuremath{\boldsymbol{#1}}}

\providecommand{\calL}{\mathcal{L}}

\providecommand{\mA}{\mat{A}}

\providecommand{\mC}{\mat{C}}
\providecommand{\mD}{\mat{D}}

\providecommand{\mF}{\mat{F}}

\providecommand{\mM}{\mat{M}}

\providecommand{\mS}{\mat{S}}

\providecommand{\mZ}{\mat{Z}}

\journal{Pattern Recognition}

\begin{document}

\begin{frontmatter}


\title{ Unsupervised Instance Segmentation with Superpixels}

\author[1]{Cuong Manh Hoang\corref{cor1}}
\ead{cuonghoang@seoultech.ac.kr}

\cortext[cor1]{Corresponding author.}
\affiliation[1]{organization={Department of Electronic Engineering, Seoul National University of Science and Technology},
            addressline={232 Gongneung-ro, Nowon-gu}, 
            city={Seoul},
            postcode={01811}, 
            country={South Korea}}

\begin{abstract}
Instance segmentation is essential for numerous computer vision applications, including robotics, human-computer interaction, and autonomous driving. Currently, popular models bring impressive performance in instance segmentation by training with a large number of human annotations, which are costly to collect. For this reason, we present a new framework that efficiently and effectively segments objects without the need for human annotations. Firstly, a MultiCut algorithm is applied to self-supervised features for coarse mask segmentation. Then, a mask filter is employed to obtain high-quality coarse masks. To train the segmentation network, we compute a novel superpixel-guided mask loss, comprising hard loss and soft loss, with high-quality coarse masks and superpixels segmented from low-level image features. Lastly, a self-training process with a new adaptive loss is proposed to improve the quality of predicted masks. We conduct experiments on public datasets in instance segmentation and object detection to demonstrate the effectiveness of the proposed framework. The results show that the proposed framework outperforms previous state-of-the-art methods.
\end{abstract}

\begin{keyword}
Instance segmentation \sep Unsupervised learning \sep Self-supervised learning
\end{keyword}

\end{frontmatter}



\section{INTRODUCTION}

Instance segmentation is a crucial task in various computer vision applications. It involves detecting individual objects within an image, providing not just bounding boxes around objects but also precise pixel-level segmentation masks for each distinct object instance. With high effectiveness, this technique proves beneficial for various areas. Specifically, it assists in recognizing and tracking objects such as pedestrians and vehicles with precise boundaries. In robotics, instance segmentation allows robots to better understand and interact with different objects in their surroundings. In addition, in agriculture, it aids in identifying and counting individual crops or plants from aerial images, facilitating monitoring of their health and growth.

\begin{figure*}[!t] \begin{center}
\begin{minipage}{0.2\linewidth}
\centerline{\includegraphics[scale=0.5]{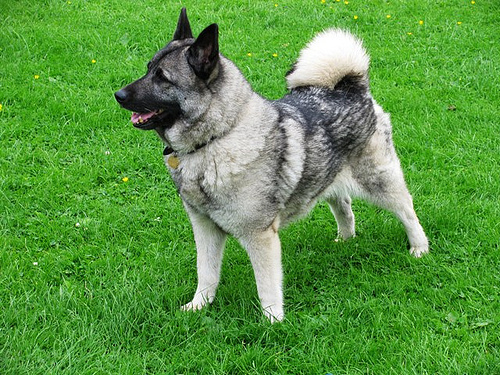}}
\end{minipage}
\begin{minipage}{0.2\linewidth}
\centerline{\includegraphics[scale=0.2]{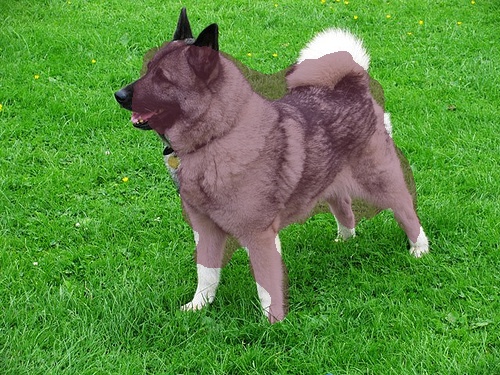}}
\end{minipage}
\begin{minipage}{0.2\linewidth}
\centerline{\includegraphics[scale=0.2]{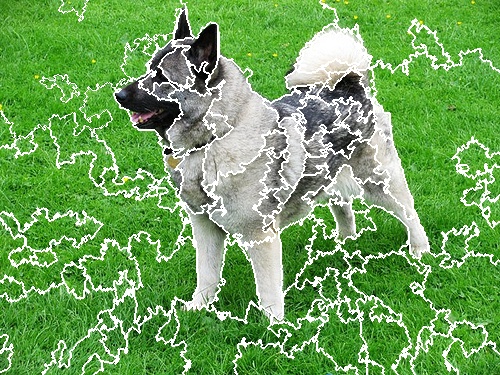}}
\end{minipage}
\begin{minipage}{0.2\linewidth}
\centerline{\includegraphics[scale=0.2]{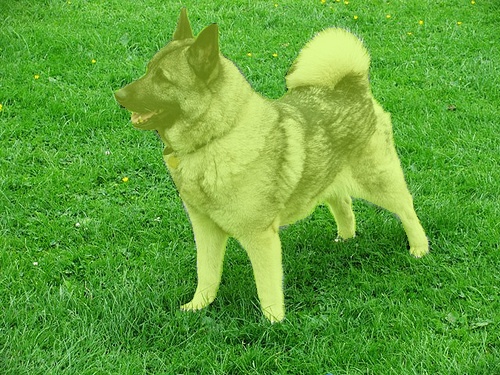}}
\end{minipage}
\\
\begin{minipage}{0.2\linewidth}
\centerline{(a)}
\end{minipage}
\begin{minipage}{0.2\linewidth}
\centerline{(b)}
\end{minipage}
\begin{minipage}{0.2\linewidth}
\centerline{(c)}
\end{minipage}
\begin{minipage}{0.2\linewidth}
\centerline{(d)}
\end{minipage}
  \caption{(a) Unlabeled image; (b) Coarse mask derived from self-supervised features; (c) Superpixels derived from low-level image features; (d) Final mask predicted by the segmentation network trained with coarse mask and superpixels.}
\label{fig:teaser}
\end{center} \end{figure*}

While supervised instance segmentation models garner significant attention and achieve impressive performance across various applications, they require a substantial amount of human annotations, which are expensive to obtain. To perform instance segmentation in an unsupervised manner, methods such as FreeSOLO \cite{wang2022freesolo} and unMORE \cite{yafeiunmore} are pretrained with in-domain unlabeled data to generate class-agnostic masks, which are subsequently used to train the segmentation network. For more practical applicability, TokenCut \cite{wang2023tokencut} introduces zero-shot unsupervised instance segmentation, which does not require in-domain unlabeled data for training. However, TokenCut \cite{wang2023tokencut} is unable to segment more than one object per image. To solve this problem, CutLER \cite{wang2023cut} presents MaskCut to segment multiple objects in an image. With significant contributions, CutLER \cite{wang2023cut} achieves impressive results in many datasets, but there are several problems that they cannot handle effectively. Firstly, their MaskCut segments a predetermined number of objects per image, which can result in under-segmentation and over-segmentation. Only exploiting these masks as supervision can degrade the performance of the model. Secondly, they improve the predicted masks through multi-round self-training, which requires a large amount of time.

To address these limitations, this paper introduces a novel framework that is able to efficiently and effectively segment class-agnostic masks of instances without any annotations. In more detail, a self-supervised Vision Transformer (ViT) initially extracts high-level image features, which are then utilized by a MultiCut algorithm to segment class-agnostic coarse masks. This algorithm can segment all possible instances in images without requiring a predefined number of objects. A mask filter is used to obtain high-quality coarse masks. Then, the high-quality coarse masks and superpixels, derived from low-level image features, are utilized to train the segmentation network using a novel loss function called superpixel-guided mask loss ($\calL_{sgm}$), which exploits pairwise affinities of color information to calculate the probability that a superpixel is foreground. This loss consists of two components. The first component, called hard loss ($\calL_{hard}$), is calculated by converting coarse masks into hard labels for superpixels. The second component, called soft loss ($\calL_{soft}$), is computed by capturing global pairwise affinities of superpixel colors to generate soft labels for superpixels. These losses are designed to integrate easily accessible forms of supervision, such as superpixels, coarse masks, and image colors, to improve performance without relying on precise human annotations. After training, the predicted masks are utilized for self-training the segmentation network using a new adaptive loss function ($\calL_{ad}$) that is based on the holistic stability of the predicted masks. This loss function further improves the quality of the predicted masks efficiently. In inference, the segmentation network can generate final high-quality masks, as shown in Figure~\ref{fig:teaser}. We evaluate the proposed framework on datasets commonly used in instance segmentation and object detection. The experimental results demonstrate that our proposed method outperforms the previous state-of-the-art methods.    

The contributions of this paper are as follows: 
\begin{itemize}
  \item We propose a novel framework that can segment objects without human annotations by leveraging self-supervised features and low-level image features.
  \item We introduce a novel superpixel-guided mask loss that leverages superpixels, along with coarse masks and color information, to improve the effectiveness of training a segmentation network.
  \item We present a self-training process with a new adaptive loss based on the holistic stability of the predicted masks to efficiently improve their quality.
  \item We demonstrate the effectiveness of the proposed framework using public datasets. 
\end{itemize}

\section{RELATED WORKS}
\subsection{Supervised Instance Segmentation}

Supervised instance segmentation is a popular task that is able to segment and classify objects in images. Current approaches for this task can be classified into various groups. Top-down methods, such as Mask R-CNN \cite{he2017mask}, detect objects and then segment them within their bounding boxes. Bottom-up methods like SSAP \cite{gao2019ssap} involve learning pixel-wise embeddings and subsequently clustering them into groups. Direct methods like SOLO \cite{wang2020solov2} segment objects without embedding learning or box detection. More recently, Mask2Former \cite{cheng2022masked} and Mask DINO \cite{li2023mask} propose strong end-to-end instance segmentation frameworks based on transformers \cite{vaswani2017attention}.

For video instance segmentation, HEVis \cite{qin2021learning} proposes a model that learns a generative probabilistic representation of instance feature distributions hierarchically and introduces normalizing flows to further improve the robustness of the learned appearance embedding. For better mask quality, HEVis* \cite{qin2023coarse} employs a holistic generative model with factorized conditional normalizing flows to capture spatio-temporal and appearance embeddings, while estimating the variance of the appearance embedding using invertible normalizing flows.

Currently, ship instance segmentation in SAR images has gained significant attention due to its valuable applications in areas such as maritime traffic regulation, fishery management, and shipwreck rescue operations. MAI-SE-Net \cite{zhang2022mask} introduces a Mask Attention Interaction and Scale Enhancement Network to improve mask-level feature interaction and enhance multiscale segmentation capabilities. HTC+ \cite{zhang2022htc+} proposes an improved Hybrid Task Cascade framework that focuses on the unique characteristics of ships in complex SAR imaging environments. FL-CSE-ROIE \cite{zhang2022full} develops a Full-Level Context Squeeze-and-Excitation ROI Extractor to better capture contextual features of the region of interest. GCBANet \cite{ke2022gcbanet} presents a Global Context Boundary-Aware Network designed to expand the receptive field and improve cross-scale bounding box prediction.

To train the instance segmentation model without the need for human annotation, our proposed model is designed to segment object masks in an unsupervised manner. It is achieved by utilizing easily obtainable forms of supervision, including coarse masks generated from self-supervised features, superpixels, and image color information.

\subsection{Unsupervised Instance Segmentation}
\begin{figure*}[t] \begin{center}
\begin{minipage}{0.49\linewidth}
\centerline{\includegraphics[scale=1.2]{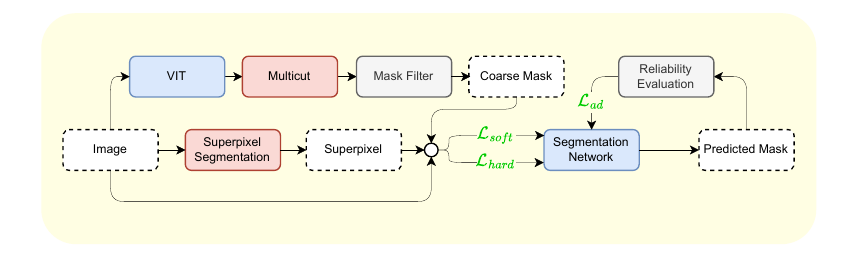}}
\end{minipage}
   \caption{Illustration of the proposed framework. High-level image features are initially extracted by a self-supervised Vision Transformer (ViT). Using these features, a MultiCut algorithm, followed by a mask filter, generates high-quality coarse masks for objects in the image. In addition to coarse masks, superpixels are segmented to serve as supervision. Then, superpixels, coarse masks, and the color information of the image are utilized to compute a novel superpixel-guided mask loss, which consists of a hard loss ($\calL_{hard}$) and a soft loss ($\calL_{soft}$), for training the segmentation network. After training, the predicted masks are exploited for self-training the segmentation network with a new adaptive loss ($\calL_{ad}$), which is based on the reliability of predicted masks. }
\label{fig:overview}
\end{center}\end{figure*}
There have been several attempts to segment masks for objects without any human annotations. COSNet \cite{lu2019see} applies a global co-attention mechanism that provides efficient and effective stages for capturing global correlations and scene context, enabling the model to focus on learning discriminative foreground representations. LOST \cite{simeoni2021localizing} and TokenCut \cite{wang2023tokencut} propose zero-shot unsupervised instance segmentation by leveraging self-supervised ViT features to segment a single salient object from each image based on a graph constructed with DINO's patch features. However, these methods are limited to detecting only one object per image. To generate multiple coarse masks per image in an unsupervised manner, FreeSOLO \cite{wang2022freesolo} introduces FreeMask, CuVLER \cite{arica2024cuvler} proposes VoteCut, and unMORE \cite{yafeiunmore} presents a two-stage pipeline comprising object-centric representation learning and multi-object reasoning. Although these techniques demonstrate impressive performance, they are pretrained on in-domain data. For zero-shot unsupervised instance segmentation, CutLER \cite{wang2023cut} proposes MaskCut to segment multiple object masks using pretrained self-supervised features, while CutS3D \cite{sick2024cuts3d} segments semantic masks in 3D to generate the final 2D instances by leveraging a point cloud representation of the scene.

In our framework, we perform zero-shot unsupervised instance segmentation using a MultiCut algorithm to segment all potential coarse masks per image, followed by a mask filter to obtain high-quality masks. We also introduce a novel superpixel-guided mask loss that leverages not only coarse masks but also other forms of supervision, such as superpixels and color information, to effectively train a segmentation network. Finally, we present a new adaptive loss to efficiently enhance the predicted masks in a self-supervised manner.

\section{PROPOSED METHOD}
\label{sec:method}

In this section, the coarse masks segmentation is firstly described. Next, we introduce a novel superpixel-guided mask loss. It combines coarse masks with superpixels and color information to train the segmentation network. Finally, a new adaptive loss is presented for self-training the segmentation network on its own predictions. The overview of the proposed framework is shown in~\fref{fig:overview}.

\subsection{Coarse Masks Generation}
\label{sec:cmg}
In unsupervised instance segmentation, coarse masks produced by an unsupervised model are utilized to train the segmentation model in place of human annotations. CutLER \cite{wang2023cut} introduced MaskCut for coarse mask segmentation. However, their approach treats the number of objects per image as a hyperparameter, which is impractical. To solve this problem, we utilize a MultiCut algorithm called RAMA \cite{abbas2022rama}, a rapid bottom-up multicut algorithm on GPU, to segment all potential objects in all images. As shown in~\fref{fig:overview}, given an image from unlabeled dataset $\mD$, a self-supervised Vision Transformer (ViT) \cite{caron2021emerging} is used to extract high-level features. These features are represented as $\mF \in \mathbb{R}^{N \times N \times E }$, where $N$ is the number of patches along one dimension and $E$ represents the feature dimension of each patch embedding. With self-supervised features, RAMA \cite{abbas2022rama} can segment all potential object masks in the image. Then, following CutLER \cite{wang2023cut}, segmented masks that contain fewer than two of the four corners are regarded as foregrounds. Because RAMA \cite{abbas2022rama} is the unsupervised segmentation technique, the quality of foreground mask is inconsistent. To address this issue, we use a mask filter to obtain high-quality object masks. With extracted features $\mF$, we compute the pairwise affinities between the patch and its 8 neighbours by cosine similarity. Then, the average pooling operation is applied to generate a pairwise affinity map $\mA \in \mathbb{R}^{N \times N }$. Given a mask $\mM \in \{0,1\}^{N \times N}$, we consider two areas: $\mM_{inner}$ containing patches within the mask and $\mM_{edge}$ containing patches around the edge of the mask. Next, the mask $\mM$ can be evaluated as follows:
\begin{equation}
	\mathcal{R}(\mM)=\frac{1}{|\mM_{inner}|}\sum_{i \in \mM_{inner}}\mA_i-\frac{1}{|\mM_{edge}|}\sum_{j \in \mM_{edge}}\mA_j
 \label{eq:fil}
\end{equation}

Where $|\mM_{inner}|$ and $|\mM_{edge}|$ represent the numbers of patches in $\mM_{inner}$ and $\mM_{edge}$. Intuitively, masks with strong inner affinities and weak edge affinities can receive a high rate. This evaluation helps to identify over- and under-segmentation issues. After that, we select the top-$Q \%$ of masks that have the highest $\mathcal{R}$ to train the segmentation network.

\subsection{Superpixel-guided Mask Loss}
\label{sec:sgm}
In this phase, we employ SOLO \cite{wang2020solov2} as our segmentation network. SOLO \cite{wang2020solov2} performs object segmentation by directly mapping an input image to the target object categories and instance masks through fully convolutional networks (FCNs), thereby bypassing the need for bounding box detection or grouping through post-processing. In this context, the segmentation network is trained with coarse masks in a class-agnostic manner. However, coarse masks, which are significantly less precise than human annotations, can negatively impact the performance of the segmentation network. For this problem, we use superpixels as an additional form of supervision to improve the mask loss. Superpixels derived from low-level image features offer several advantages, including excellent preservation of object boundaries, flexibility, reduced noise, and lower computational complexity. Therefore, it is a promising form of supervision that can be employed to train the segmentation network. While any algorithm can be used, we employ Multiscale Combinatorial Grouping (MCG) \cite{arbelaez2014multiscale} to generate high-quality superpixels. 

To use superpixels in training, we initially segment superpixels $\mS$, and the color vector of a superpixel $\mS_k$ is computed as follows:
\begin{equation}
	\mu_k=\frac{1}{|\mS_k|}\sum_{i \in \mS_k}\mC_i
 \label{eq:avgsup}
\end{equation}
where $|\mS_k|$ denotes number of pixels within the superpixel $\mS_k$. $\mC_i \in \mathbb{R}^{3}$ is a color vector of a pixel belonging to superpixel $\mS_k$. Next, the color similarities between a pixel and its corresponding superpixel are computed as follows:
\begin{equation}
	\delta_{k,i}=\text{exp}(-\frac{\norm{\mu_k-\mC_i}^2_2}{\alpha_1})
 \label{eq:colorsim}
\end{equation}
where $\alpha_1$ is a hyperparameter.
Given a probablity map predicted by the network $\widetilde{\mM} \in (0,1)^{H \times W}$, the probability of superpixel $S_k$ being foreground can be calculated as follows:
\begin{equation}
	P_k=\frac{1}{\upsilon_k}\sum_{i \in \mS_k}\widetilde{\mM}_i.\delta_{k,i} \;\text{,} \; \upsilon_k=\sum_{i \in \mS_k}\delta_{k,i}
 \label{eq:prosup}
\end{equation}
where $P_k$ is largely influenced by the probabilities of pixels whose color vectors are similar to $\mu_k$. Therefore, these pixels receive a larger gradient during training, while noisy pixels receive a smaller gradient. Then, we define $y_k \in \{0,1\}$ as the label for the superpixel $\mS_k$. Since all the pixels within a superpixel share similar color, texture, or other low-level features, they are highly to belong to the same object. Given a coarse mask $\mM$, we assume that $y_k=1$ when all its pixels are classified as foreground, and $y_k=0$ when all its pixels are classified as background. After labeling, superpixels that contain both foreground and background pixels are considered unlabeled and are ignored during training. Then, the hard loss is computed as follows:  
\begin{equation}
	\calL_{hard}=-\frac{1}{N_s}\sum_{k=1}^{N_s}(y_k.\text{log}(P_k)+(1-y_k).\text{log}(1-P_k))
 \label{eq:term1}
\end{equation}
where $N_s$ is the number of labeled superpixels. By leveraging the benefits of superpixels, coarse masks, and color vectors, this loss function makes the model more robust to noise and significantly improves its performance.

\begin{figure*}[!t] \begin{center}
\begin{minipage}{0.49\linewidth}
\centerline{\includegraphics[scale=1.2]{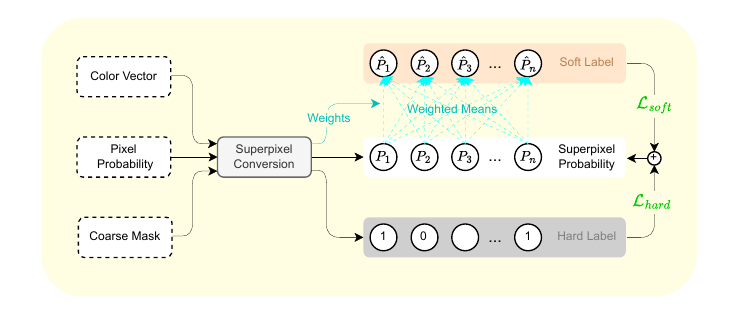}}
\end{minipage}
   \caption{Illustration of the superpixel-guided mask loss ($\calL_{sgm}$). Given segmented superpixels, the pixel probabilities predicted by the segmentation network and the color vectors are used to compute superpixel probabilities. For the training process, coarse masks are converted into hard labels for superpixels used to calculate hard loss ($\calL_{hard}$). In this figure, the empty regions represent unlabeled superpixels that are ignored in the calculation of the hard loss. Apart from hard labels, superpixel probabilities are used to compute soft labels through global propagation in the form of weighted means. Here, the weights are global pairwise affinities computed from the color vectors of the superpixels. Then, the soft labels are used to compute soft loss ($\calL_{soft}$), which is summed with $\calL_{hard}$ to form $\calL_{sgm}$. }
\label{fig:sgm}
\end{center}\end{figure*}

In the hard loss ($\calL_{hard}$), the gradient received by a pixel is influenced by the local pixels within its superpixel. However, local operations are unable to capture global context cues and long-range affinity dependencies. Moreover, the hard loss depends on coarse masks, which often contain noise, resulting in unlabeled superpixels. For this reason, we propose a soft loss ($\calL_{soft}$), which captures global pairwise affinities to generate soft labels for all superpixels. Specifically, we build a graph $\mathcal{G}$ with superpixels as nodes of the graph. A superpixel is connected to all of its adjacent superpixels. The weight of an edge is measured by similarity between the color vectors of two adjacent superpixels, $i.e, w_{m,n}=\norm{\mu_m-\mu_n}^2_2$. Inspired by the tree-based method \cite{song2019learnable}, we apply the minimum spanning tree (MST) algorithm to remove edges with large weights, resulting in a tree-based sparse graph $\mathcal{G}_T$. In this graph, a node can propagate long-range affinities to other nodes. Although distant nodes along the spanning tree must pass through nearby nodes on the path, the distance-insensitive maximum affinity function can ease this geometric constraint and reduce the affinity decay for long-range nodes. Therefore, the global pairwise function is computed as follows:
\begin{equation}
	\psi_{k,l}=\text{exp}(-\max_{\forall(m,n) \in \mathbb{E}_{k,l}} \frac{w_{m,n}}{\alpha_2})
 \label{eq:globalaffi}
\end{equation}
where $\mathbb{E}_{k,l}$ is the set of edges in the path from node $k$ to node $l$. $w_{m,n}$ denotes edge weight between the adjacent nodes $m$ and $n$. $\alpha_2$ is a hyperparameter. Then, the probability of superpixel $\mS_k$ being foreground can be computed by the probabilities of other superpixels as follows:
\begin{equation}
	\hat{P}_k=\frac{1}{\gamma_k}\sum_{l \in \mS}P_l.\psi_{k,l} \;\text{,} \; \gamma_k=\sum_{l \in \mS}\psi_{k,l}
 \label{eq:globalprob}
\end{equation}
The generated $\hat{P}_k$ can be used as soft label to train the model with $L_1$ distance as follows:
\begin{equation}
	\calL_{soft}=\frac{1}{|\mS|}\sum_{k \in \mS}\norm{P_k-\hat{P}_k}_1
 \label{eq:term2}
\end{equation}
where $|\mS|$ is the number of all superpixels. Using this loss, the probability of a pixel is calculated by the probabilities of other pixels to capture global contexts, and processing through superpixels can reduce the negative effects of global noise. Moreover, constructing a graph based on superpixels rather than individual pixels can decrease the number of nodes and edges, hence significantly reducing computational complexity. Lastly, the superpixel-guided mask loss is computed as follows:
\begin{equation}
	\calL_{sgm}=\calL_{hard}+\calL_{soft}
 \label{eq:final}
\end{equation}

The illustration of superpixel-guided mask loss is shown in \fref{fig:sgm}. 

\subsection{Adaptive Loss for Self-Training}
\label{sec:adap}
After training, the predicted masks are considerably better than the original coarse masks \cite{wang2023cut,wang2022freesolo}. Hence, we propose a new adaptive loss for self-training to improve performance of the segmentation network. Inspired by \cite{yang2022st++}, we use holistic stability to evaluate the reliability of the predicted masks. We store $e$ checkpoints of the segmentation model. Each checkpoint is saved after a fixed number of epochs. Given a mask predicted by the last checkpoint $\mathbf{m}_i^e$, we compute the score for it as follows:
\begin{equation}
	\mZ_i=\sum_{j=1}^{e-1}\text{IoU}(\mathbf{m}^e_i, \mathbf{m}^j_i)
 \label{eq:scorez}
\end{equation}
where $\mathbf{m}^j_i$ is a mask generated by an intermediate checkpoint that has the highest IoU overlap with $\mathbf{m}_i^e$. Intuitively, the predicted mask $\mathbf{m}^e_i$ is reliable when its score $\mZ_i$ is high. Then, we collect scores of all masks in a list $\mZ$ and implement Min-Max normalization as follows:
\begin{equation}
	\bar{\mZ}_i=\frac{\mZ_i-\text{min}(\mZ)}{\text{max}(\mZ)-\text{min}(\mZ)}(1-\epsilon)+\epsilon \;\text{,} \; 
 \label{eq:scorenorm}
\end{equation}
where $\bar{\mZ}_i \in [\epsilon,1]$ and $\epsilon \in (0,1)$ is a hyperparameter. After normalization, $\bar{\mZ}_i$ can be used to weight the loss function. Considering the predicted mask $\mathbf{m}^e_i$ that is used to train the model, its boundary region is likely to be noisy because it usually corresponds to the decision boundary \cite{wang2022noisy}. Therefore, we propose a weight for a pixel position as follows:
\begin{equation}
	\begin{split}
	\varphi_j=  \left\{
	\begin{array}{l l}
	\bar{\mZ}_i, \; \text{ if } d_{j} \leq \hat{d} \\
	1, \; \text{ otherwise } \\ 
	\end{array} \right.
\end{split}
\end{equation}
where $d_j$ denotes the Euclidean distance from the pixel to its nearest object boundary, and $\hat{d}$ represents a hyperparameter determining the boundary regions. Lastly, the adaptive loss is computed as follows:
\begin{equation}
	\calL_{ad}=-\frac{1}{N_p}\sum_{j=1}^{N_p}\varphi_j.(y_j.\text{log}(p_j)+(1-y_j).\text{log}(1-p_j))
 \label{eq:adap}
\end{equation}
where $p_j$ is the predicted probability. $y_j$ is the label from $\mathbf{m}_i^e$. $N_p$ is the number of pixels. The self-training with adaptive loss efficiently leverages reliable labels and reduces the negative effects of unreliable labels to enhance performance.

\section{EXPERIMENTS AND RESULTS}
\label{sec:result}

\subsection{Experimental Setting}

\begin{table*}[!ht]
\footnotesize
\begin{center}

\caption{Summary of datasets used for unsupervised zero-shot evaluation.}
\label{tab:data}
\begin{tabular}{>{\centering}m{0.19\textwidth}>{\centering}m{0.19\textwidth}>{\centering}m{0.19\textwidth}>{\centering\arraybackslash}m{0.19\textwidth}} 
 \hline
Dataset & Testing Set & $\#$images & Mask Label \\ 
\hline\hline
COCO \cite{lin2014microsoft} & val2017 & 5000 & \ding{51} \\ 
COCO20K \cite{simeoni2021localizing} & COCO subset & 19817 & \ding{51} \\ 
Pascal VOC~\citep{everingham2010pascal} & trainval07 & 9963 & \ding{55} \\ 
UVO~\citep{wang2021unidentified} & val & 7356 & \ding{51} \\ 
KITTI~\citep{geiger2012we} & trainval & 7521 & \ding{55} \\ 

\hline
\end{tabular}

\end{center}
\end{table*}

\begin{table*}[!t]
\scriptsize
\begin{center}
\caption{Unsupervised object detection and instance segmentation on COCO 20K \cite{simeoni2021localizing} and COCO val2017 \cite{lin2014microsoft}. *: methods train on extra unlabeled images from the downstream datasets, while other methods only train on ImageNet \cite{deng2009imagenet}.}
\label{tab:result_coco}
\begin{tabular}{>{\raggedright}m{0.15\textwidth}|>{\centering}m{0.034\textwidth}>{\centering}m{0.034\textwidth}>{\centering\arraybackslash}m{0.034\textwidth}>{\centering\arraybackslash}m{0.034\textwidth}>{\centering\arraybackslash}m{0.034\textwidth}>{\centering\arraybackslash}m{0.034\textwidth}|>{\centering\arraybackslash}m{0.034\textwidth}>{\centering\arraybackslash}m{0.034\textwidth}>{\centering\arraybackslash}m{0.034\textwidth}>{\centering\arraybackslash}m{0.034\textwidth}>{\centering\arraybackslash}m{0.034\textwidth}>{\centering\arraybackslash}m{0.034\textwidth}}  
\hline
\multirow{2}{*}{Methods} & \multicolumn{6}{c|}{COCO 20K} & \multicolumn{6}{c}{COCO val2017} \\
\cline{2-13}
 & $\text{AP}^{\text{box}}_{\text{50}}$ & $\text{AP}^{\text{box}}_{\text{75}}$ & $\text{AP}^{\text{box}}$ & $\text{AP}^{\text{mask}}_{\text{50}}$ & $\text{AP}^{\text{mask}}_{\text{75}}$ & $\text{AP}^{\text{mask}}$ & $\text{AP}^{\text{box}}_{\text{50}}$ & $\text{AP}^{\text{box}}_{\text{75}}$ & $\text{AP}^{\text{box}}$ & $\text{AP}^{\text{mask}}_{\text{50}}$ & $\text{AP}^{\text{mask}}_{\text{75}}$ & $\text{AP}^{\text{mask}}$ \\
\hline\hline
$\text{LOST}^*$ \cite{simeoni2021localizing}& - & - & - & 2.4 & 1.0 & 1.1 & - & - & - & - & - & -  \\ 
DINO \cite{caron2021emerging}& 1.7 & 0.1 & 0.3 & - & - & -& - & - & - & -& - & -  \\ 
$\text{MaskDistill}^*$ \cite{van2022discovering}& - & - & - & 6.8 & 2.1 & 2.9 & - & - & - & - & - & -   \\ 
TokenCut \cite{wang2023tokencut}& - & - & - & -& - & -& 5.8 & 2.8 & 3.0 & 4.8& 1.9 & 2.4  \\ 
$\text{FreeSOLO}^*$ \cite{wang2022freesolo}& 9.7 & 3.2 & 4.1 & 9.7 & 3.4 & 4.3 & 9.6 & 3.1 & 4.2 & 9.4 & 3.3 & 4.3  \\ 
CutLER \cite{wang2023cut}& 22.4 & 11.9 & 12.5 & 19.6 & 9.2 & 10.0 & 21.9 & 11.8 & 12.3 & 18.9 & 9.2 & 9.7  \\ 
$\text{CuVLER}^*$ \cite{arica2024cuvler}& 24.1 & 12.3 & 13.1 & 21.6 & 9.7 & 10.7& 23.5 & 12.0 & 12.8 & 20.4 & 9.6 & 10.4 \\ 
CutS3D \cite{sick2024cuts3d}& 24.6 & 12.5 & 13.4 & 21.3 & 9.9 & 10.9& 24.3 & 12.5 & 13.3 & 20.8 & 9.8 & 10.7 \\ 
$\text{unMORE}^*$ \cite{yafeiunmore}& 25.9 & 13.0 & 13.9 & 23.6 & 11.1 & 12.0 & 25.4 & 12.7 & 13.6 & 22.9 & 10.7 &11.7\\ 
Ours& \textbf{35.3} & \textbf{17.2} & \textbf{18.8} & \textbf{29.1}& \textbf{14.3} &\textbf{14.8} & \textbf{34.5} & \textbf{16.7} & \textbf{18.4} & \textbf{28.6}& \textbf{13.8}& \textbf{14.6} \\ 

 \hline
\end{tabular}
\end{center}
\end{table*}

\begin{table*}[t]
\footnotesize
\begin{center}
\caption{Zero-shot unsupervised object detection on Pascal VOC dataset~\citep{everingham2010pascal}.}
\label{tab:voc}
\begin{tabular}{>{\raggedright}m{0.2\textwidth}|>{\centering}m{0.067\textwidth}>{\centering}m{0.067\textwidth}>{\centering\arraybackslash}m{0.067\textwidth}>{\centering\arraybackslash}m{0.067\textwidth}>{\centering\arraybackslash}m{0.067\textwidth}>{\centering\arraybackslash}m{0.067\textwidth}} 
 \hline
Methods & $\text{AP}_{\text{50}}$ & $\text{AP}_{\text{75}}$ & $\text{AP}$ & $\text{AP}_{\text{S}}$ & $\text{AP}_{\text{M}}$ & $\text{AP}_{\text{L}}$\\ 
\hline\hline
LOST \cite{simeoni2021localizing} &  19.8 & - & 6.7 & -&-&- \\ 
FreeSOLO~\citep{wang2022freesolo} & 15.9 & 3.6 & 5.9 & 0.0&2.0&9.3 \\ 
CutLER~\citep{wang2023cut} & 36.9 & 19.2 & 20.2 & 1.3 & 6.5 & 32.2 \\ 
CuVLER \cite{arica2024cuvler}& 39.4 & 19.4 & 22.3 & 1.4& 7.2 & 32.6 \\ 
CutS3D \cite{sick2024cuts3d}& 40.8 & 19.8 & 21.4 & 1.2 & 7.6 & 33.7 \\ 
unMORE \cite{yafeiunmore}& 40.4 & 21.5 & 22.7 & 2.1 & 8.4 & 34.9 \\ 
Ours & \textbf{42.8} & \textbf{28.7} & \textbf{33.1}& \textbf{3.2} & \textbf{11.6} & \textbf{39.2} \\ 
\hline
\end{tabular}
\end{center}
\end{table*}

\begin{table*}[!t]
\centering 
\footnotesize
\caption{Zero-shot unsupervised object detection and instance
segmentation on the UVO dataset~\citep{wang2021unidentified}.}
\label{tab:uvo}
\begin{tabular}{>{\raggedright}m{0.2\textwidth}|>{\centering}m{0.067\textwidth}>{\centering}m{0.067\textwidth}>{\centering\arraybackslash}m{0.07\textwidth}>{\centering\arraybackslash}m{0.067\textwidth}>{\centering\arraybackslash}m{0.067\textwidth}>{\centering\arraybackslash}m{0.067\textwidth}}
 \hline
Methods & $\text{AP}^{\text{box}}_{\text{50}}$ & $\text{AP}^{\text{box}}_{\text{75}}$ & $\text{AP}^{\text{box}}$ & $\text{AP}^{\text{mask}}_{\text{50}}$ & $\text{AP}^{\text{mask}}_{\text{75}}$ & $\text{AP}^{\text{mask}}$\\ 
\hline\hline
MaskDistill \cite{van2022discovering}& 7.3 & 1.5 & 2.2 & 5.8 & 1.1 & 1.7   \\ 
TokenCut \cite{wang2023tokencut} & 8.7 & 1.3 & 2.4 & 7.5& 0.8&1.5 \\ 
FreeSOLO~\citep{wang2022freesolo} & 10.0 & 1.8 & 3.2 &9.5&2.0&3.3 \\ 
CutLER~\citep{wang2023cut} & 31.7 & 14.1 & 16.1 & 22.8 & 8.0 & 10.1 \\ 
CutS3D \cite{sick2024cuts3d}& 32.1 & 17.3 & 19.6 & 24.7 & 9.1 & 11.5 \\ 
unMORE \cite{yafeiunmore}& 33.2 & 18.6 & 21.5 & 26.3 & 9.7 & 12.3 \\ 
Ours & \textbf{38.1} & \textbf{23.8} & \textbf{29.3}& \textbf{33.4} & \textbf{11.9} & \textbf{15.4} \\ 
\hline
\end{tabular}

\end{table*}

\begin{table*}[t!]
\footnotesize
\begin{center}
\caption{Zero-shot unsupervised object detection on KITTI dataset~\citep{geiger2012we}.}
\label{tab:kitti}
\begin{tabular}{>{\raggedright}m{0.2\textwidth}|>{\centering}m{0.067\textwidth}>{\centering}m{0.067\textwidth}>{\centering\arraybackslash}m{0.07\textwidth}>{\centering\arraybackslash}m{0.067\textwidth}>{\centering\arraybackslash}m{0.067\textwidth}>{\centering\arraybackslash}m{0.067\textwidth}}
 \hline
 Methods &$\text{AP}_{\text{50}}$ & $\text{AP}_{\text{75}}$ & $\text{AP}$ & $\text{AP}_{\text{S}}$ & $\text{AP}_{\text{M}}$ & $\text{AP}_{\text{L}}$ \\ 
 \hline \hline
  FreeSOLO~\citep{wang2022freesolo} & 9.1 & 2.5 & 3.9 & 0.0 &1.8&8.2 \\
CutLER~\citep{wang2023cut} & 18.4 &6.7 &8.5& 0.5& 5.6& 19.2 \\ 

CutS3D \cite{sick2024cuts3d}& 21.1 & 7.6 & 9.7 & 0.1 & 5.7 & 22.9 \\ 
unMORE \cite{yafeiunmore}& 26.7 & 12.6 & 13.7 & 1.9 & 9.7 & 22.8 \\ 
 Ours & \textbf{28.5} & \textbf{13.9} & \textbf{16.3} & \textbf{2.8}& \textbf{11.1}& \textbf{26.5} \\ 
 \hline
\end{tabular}
\end{center}
\end{table*}

\noindent \textbf{Dataset.} Following CutLER \cite{wang2023cut}, we only use ImageNet \cite{deng2009imagenet} (1.3 million images) for training and do not use any type of annotations. Our model is subsequently evaluated in a zero-shot manner on public datasets, including COCO val2017 \cite{lin2014microsoft}, COCO20K \cite{simeoni2021localizing}, PASCAL VOC trainval07 \cite{everingham2010pascal}, UVO val \cite{wang2021unidentified}, KITTI trainval~\citep{geiger2012we}. COCO \cite{lin2014microsoft} is a large-scale object detection and instance segmentation dataset, containing about 115K and 5K images in the training and validation split, respectively. We evaluate our model in a class-agnostic manner on COCO val2017 and COCO 20K, which is a subset of the COCO trainval2014 \cite{lin2014microsoft}. COCO 20K includes 19817 randomly sampled images and is used as a benchmark in \cite{simeoni2021localizing,wang2023tokencut}. Pascal VOC \cite{everingham2010pascal} is another popular benchmark for object detection. We test our model on its trainval07 split in COCO style evaluation metrics.  Unidentified Video Objects (UVO) \cite{wang2021unidentified} is an exhaustively annotated dataset for video object detection and instance segmentation. We evaluate our model on UVO val by frame-by-frame inference and report results in COCO style evaluation metrics. KITTI~\citep{geiger2012we}, captured from vehicle-mounted cameras in real-world driving scenarios, is used to evaluate our model on 7521 images from the trainval split. \tref{tab:data} shows the summary of these datasets.

\noindent \textbf{Implementation Details.} In coarse mask generation, we use ViT-B/8 \cite{dosovitskiy2020image} DINO \cite{caron2021emerging} model to extract self-supervised features. The hyperparameter $Q=60 \%$. For segmentation network, we use SOLO model \cite{wang2020solov2} with  ResNet-101 \cite{he2016deep} as the backbone. For superpixel-guided mask loss, we set hyperparameters $\alpha_1=100$, $\alpha_2=200$. For adaptive loss, we set hyperparameters $e=3, \epsilon=0.6$, $\hat{d}=3$. The experiments were conducted on a computer with two Nvidia GeForce RTX 3090 GPUs, an Intel Core i9-10940X CPU, and 128 GB RAM.

\noindent \textbf{Evaluation Metric.} Following CutLER \cite{wang2023cut}, we report class-agnostic COCO style averaged precision for object detection and segmentation tasks. The evaluation metrics include $\text{AP}_{50}$ at a fixed IoU threshold of 0.5, $\text{AP}_{75}$ at a fixed IoU threshold of 0.75 and AP across multiple IoU thresholds from 0.5 to 0.95, with a step size of 0.05.  
\subsection{Result}
To demonstrate the effectiveness of our proposed framework, we compare it with existing methods on various datasets.

\tref{tab:result_coco} shows quantitative results on COCO 20K \cite{simeoni2021localizing} and COCO val2017 \cite{lin2014microsoft}. We evaluate the models by using box and mask annotations.  In this setting, DINO \cite{caron2021emerging}, TokenCut \cite{wang2023tokencut}, CutLER \cite{wang2023cut}, CutS3D \cite{sick2024cuts3d} and ours perform zero-shot unsupervised object detection and segmentation by only train on ImageNet \cite{deng2009imagenet}, while LOST \cite{simeoni2021localizing}, MaskDistill \cite{van2022discovering}, FreeSOLO \cite{wang2022freesolo}, CuVLER \cite{arica2024cuvler} and unMORE \cite{yafeiunmore} require additional training on extra unlabeled images from downstream datasets. The results show that our model overcomes most of the existing frameworks. For COCO 20K dataset \cite{simeoni2021localizing}, our model achieves the best scores with 18.8$\%$ in $\text{AP}^{\text{box}}$ and 14.8$\%$ in $\text{AP}^{\text{mask}}$. unMORE \cite{yafeiunmore} achieves the second-best scores with 13.9$\%$ in $\text{AP}^{\text{box}}$ and 12.0$\%$ in $\text{AP}^{\text{mask}}$. For COCO val2017 \cite{lin2014microsoft}, our model surpasses unMORE \cite{yafeiunmore} by 4.8$\%$ in $\text{AP}^{\text{box}}$ and 2.9$\%$ in $\text{AP}^{\text{mask}}$. This superiority is mainly due to using superpixels along with coarse masks to enhance the model's performance.

\begin{table*}[!t]
\footnotesize
\begin{center}
\caption{Unsupervised object detection and instance
segmentation on SSDD dataset~\citep{zhang2021sar}.}
\label{tab:sar}
\begin{tabular}{>{\raggedright}m{0.2\textwidth}|>{\centering}m{0.067\textwidth}>{\centering}m{0.067\textwidth}>{\centering\arraybackslash}m{0.067\textwidth}>{\centering\arraybackslash}m{0.067\textwidth}>{\centering\arraybackslash}m{0.067\textwidth}>{\centering\arraybackslash}m{0.067\textwidth}} 
 \hline
 Methods &$\text{AP}^{\text{box}}_{\text{50}}$ & $\text{AP}^{\text{box}}_{\text{75}}$ & $\text{AP}^{\text{box}}$ & $\text{AP}^{\text{mask}}_{\text{50}}$ & $\text{AP}^{\text{mask}}_{\text{75}}$ & $\text{AP}^{\text{mask}}$  \\ 
 \hline \hline
  FreeSOLO~\citep{wang2022freesolo} & 8.6 & 2.0 & 3.7 &7.2&1.6&3.0 \\ 
CutLER~\citep{wang2023cut} & 17.1 & 5.9 & 8.0 & 16.5 & 5.4 & 7.2 \\ 
CutS3D \cite{sick2024cuts3d}& 18.9 & 7.2 & 9.1 & 17.4 & 5.8 & 8.0\\
unMORE \cite{yafeiunmore}& 20.2 & 8.7 & 11.4 & 18.0 & 6.5 & 8.6 \\ 
 Ours & \textbf{27.5} & \textbf{13.4} & \textbf{15.8} & \textbf{21.8}& \textbf{8.2}& \textbf{11.9} \\ 
 \hline
\end{tabular}
\end{center}
\end{table*}

\begin{figure*}[!t] 
\begin{center}



\begin{minipage}{0.03\linewidth}
\centerline{(a)}
\end{minipage}
\begin{minipage}{0.2\linewidth}
\centerline{\includegraphics[scale=0.2]{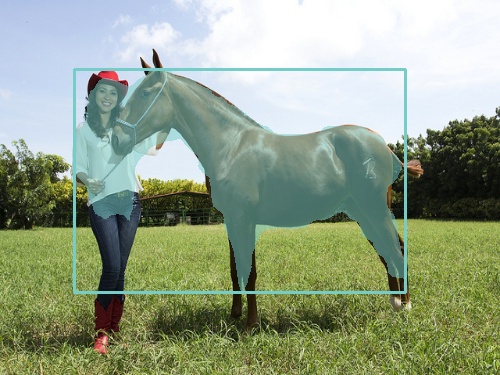}}
\end{minipage}
\begin{minipage}{0.2\linewidth}
\centerline{\includegraphics[scale=0.2]{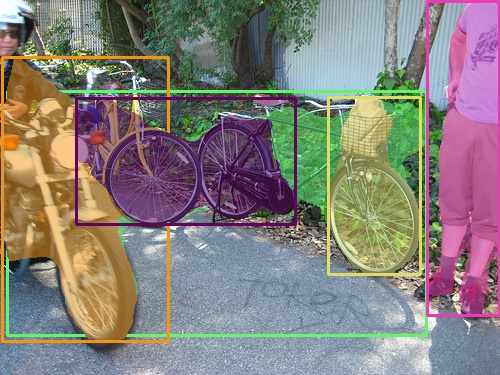}}
\end{minipage}
\begin{minipage}{0.2\linewidth}
\centerline{\includegraphics[scale=0.2]{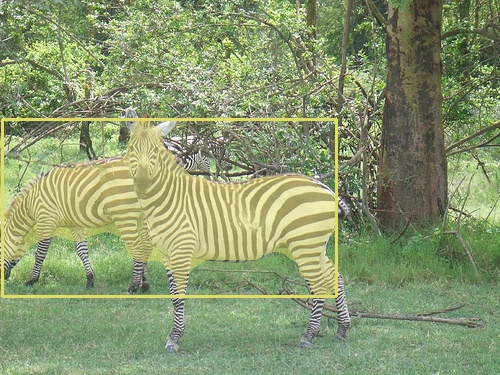}}
\end{minipage}
\begin{minipage}{0.2\linewidth}
\centerline{\includegraphics[scale=0.2]{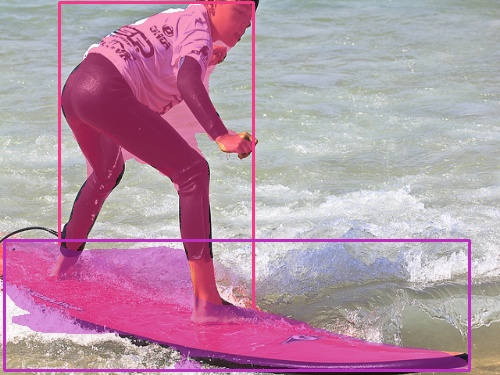}}
\end{minipage}
\\
\vspace{0.1cm}
\begin{minipage}{0.03\linewidth}
\centerline{(b)}
\end{minipage}
\begin{minipage}{0.2\linewidth}
\centerline{\includegraphics[scale=0.2]{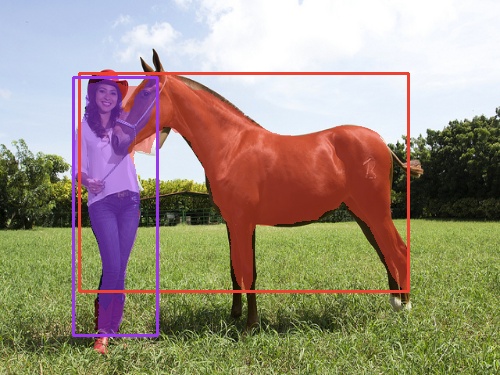}}
\end{minipage}
\begin{minipage}{0.2\linewidth}
\centerline{\includegraphics[scale=0.2]{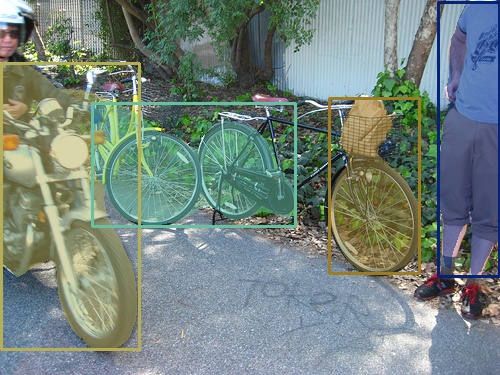}}
\end{minipage}
\begin{minipage}{0.2\linewidth}
\centerline{\includegraphics[scale=0.2]{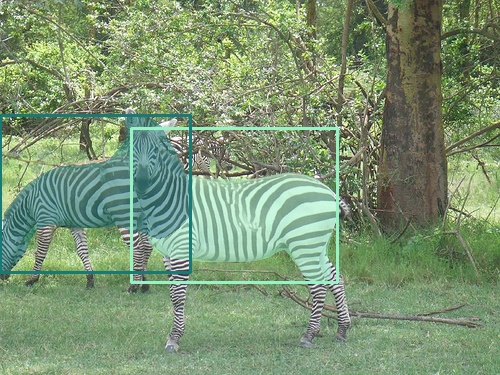}}
\end{minipage}
\begin{minipage}{0.2\linewidth}
\centerline{\includegraphics[scale=0.2]{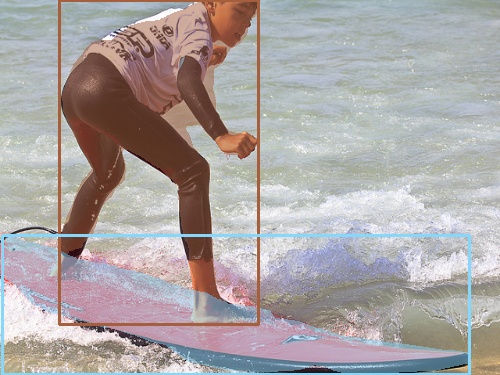}}
\end{minipage}
\\
\vspace{0.1cm}
\begin{minipage}{0.03\linewidth}
\centerline{(c)}
\end{minipage}
\begin{minipage}{0.2\linewidth}
\centerline{\includegraphics[scale=0.2]{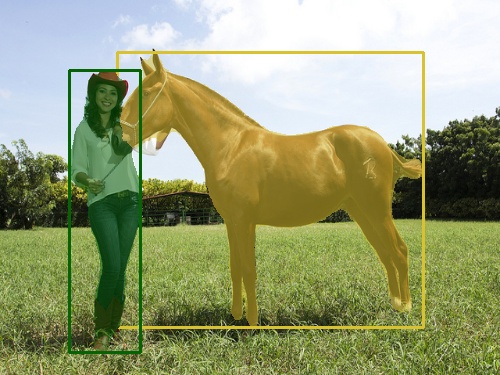}}
\end{minipage}
\begin{minipage}{0.2\linewidth}
\centerline{\includegraphics[scale=0.2]{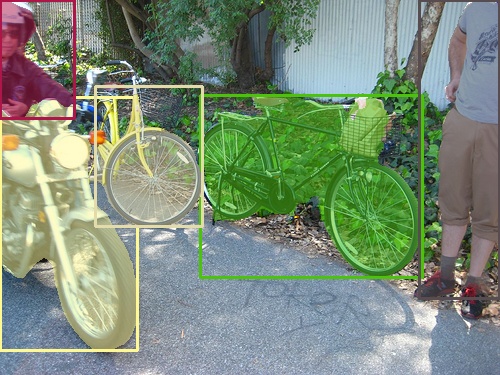}}
\end{minipage}
\begin{minipage}{0.2\linewidth}
\centerline{\includegraphics[scale=0.2]{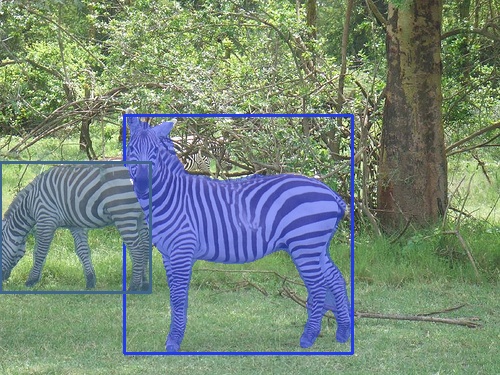}}
\end{minipage}
\begin{minipage}{0.2\linewidth}
\centerline{\includegraphics[scale=0.2]{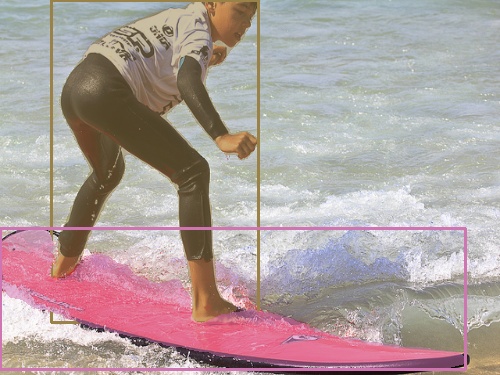}}
\end{minipage}

\caption{Qualitative results on the COCO val2017 \cite{lin2014microsoft}. (a) CutLER \cite{wang2023cut}; (b) unMORE \cite{yafeiunmore}; (c) Ours.}
\label{fig:result_cocoval}
\end{center}
\end{figure*}

Quantitative results on the PASCAL VOC \cite{everingham2010pascal} are shown in~\tref{tab:voc}. The metrics are computed using box annotations. All models perform zero-shot unsupervised object detection without training on downstream datasets. The proposed method is compared to LOST \cite{simeoni2021localizing}, FreeSOLO \cite{wang2022freesolo}, CutLER \cite{,wang2023cut}, CuVLER \cite{arica2024cuvler}, CutS3D \cite{sick2024cuts3d} and unMORE \cite{yafeiunmore}. The results show that our model exceeds the state of the art in all metrics, with significant increases of 7.2$\%$ in $\text{AP}_{\text{75}}$ and 10.4$\%$ in AP.

\tref{tab:uvo} shows the results on UVO dataset~\citep{wang2021unidentified}. The models are evaluated by using box and mask annotations. All models perform zero-shot unsupervised object detection and instance segmentation without training on downstream datasets. The proposed method is compared to MaskDistill \cite{van2022discovering}, TokenCut \cite{wang2023tokencut}, FeeSOLO \cite{wang2022freesolo}, CutLER \cite{wang2023cut}, CutS3D \cite{sick2024cuts3d} and unMORE \cite{yafeiunmore}. The results show that our model achieves the best performance with 29.3$\%$ in $\text{AP}^{\text{box}}$ and 15.4$\%$ in $\text{AP}^{\text{mask}}$, while unMORE \cite{yafeiunmore} achieves the second-best scores with 21.5$\%$ in $\text{AP}^{\text{box}}$ and 12.3$\%$ in $\text{AP}^{\text{mask}}$.

\tref{tab:kitti} presents the quantitative results on the KITTI dataset \cite{geiger2012we}, which includes real-world driving scenes with diverse lighting, motion blur, occlusion, and weather conditions. As a result, evaluating the performance on this dataset can demonstrate the effectiveness of our model on a lower-quality dataset. The proposed method is compared to FreeSOLO \cite{wang2022freesolo}, CutLER \cite{wang2023cut}, CutS3D \cite{sick2024cuts3d} and unMORE \cite{yafeiunmore}. The table shows that our model outperforms the state-of-the-art unMORE \cite{yafeiunmore}, with significant improvements of 2.6$\%$ in AP and 3.7$\%$ in $\text{AP}_{\text{L}}$. These results suggest that our model brings superior performance even with low-quality images.

\tref{tab:sar} shows the performance of our model in unsupervised SAR ship instance segmentation, demonstrating the generalizability and flexibility of our model in a different application scenario. Currently, research about SAR images brings lots of advantages in real applications. G-CNN \cite{zhang2019high} and DS-CNN \cite{zhang2019depthwise} propose models based on grid convolutional neural networks and depthwise separable convolutional neural networks, respectively, for high-speed SAR ship detection. PFGFE-Net \cite{zhang2022polarization} introduces a polarization fusion network with geometric feature embedding for SAR ship classification. With our model, we perform unsupervised SAR ship instance segmentation, which segments ship instances without training on human annotations. For the experiment, we use the SSDD dataset \cite{zhang2021sar}, which includes 1160 SAR images. The training-test ratio is 4:1. For implementation, we first follow G-CNN \cite{zhang2019high} to apply preprocessing for data enhancement, and then train DINO \cite{caron2021emerging} and our segmentation network on the training set in an unsupervised manner. Subsequently, the test set is used for evaluation. We also implement FreeSOLO~\citep{wang2022freesolo}, CutLER~\citep{wang2023cut}, CutS3D \cite{sick2024cuts3d} and unMORE \cite{yafeiunmore} in the same manner for comparison. \tref{tab:sar} shows that our model achieves the best performance, with 15.8$\%$ in $\text{AP}^{\text{box}}$ and 11.9$\%$ in $\text{AP}^{\text{mask}}$, respectively. These results indicate that our model offers better capability to generalize and perform flexibly across applications.

\begin{figure*}[!t] 
\begin{center}
\begin{minipage}{0.03\linewidth}
\centerline{(a)}
\end{minipage}
\begin{minipage}{0.2\linewidth}
\centerline{\includegraphics[scale=0.2]{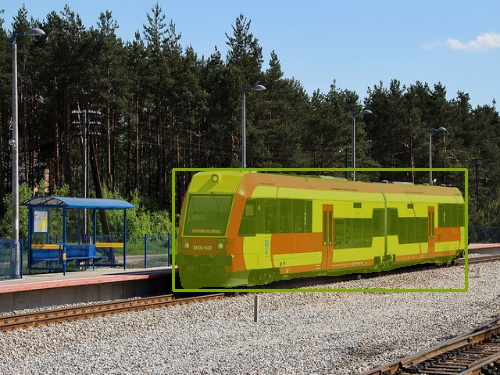}}
\end{minipage}
\begin{minipage}{0.2\linewidth}
\centerline{\includegraphics[scale=0.2]{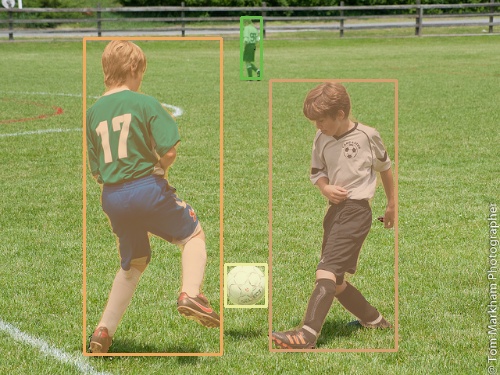}}
\end{minipage}
\begin{minipage}{0.2\linewidth}
\centerline{\includegraphics[scale=0.2]{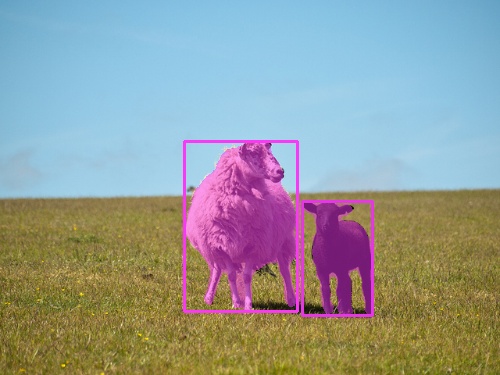}}
\end{minipage}
\begin{minipage}{0.2\linewidth}
\centerline{\includegraphics[scale=0.2]{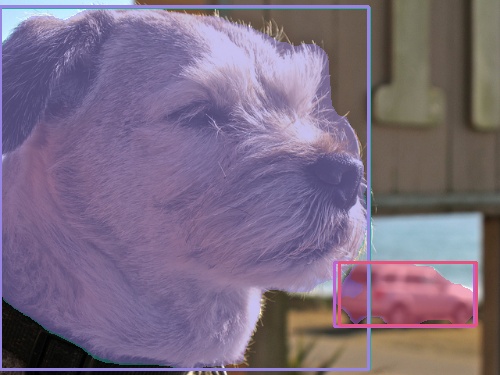}}
\end{minipage}
\\
\vspace{0.1cm}
\begin{minipage}{0.03\linewidth}
\centerline{(b)}
\end{minipage}
\begin{minipage}{0.2\linewidth}
\centerline{\includegraphics[scale=0.2]{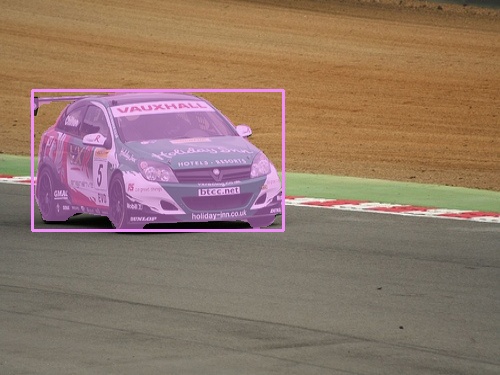}}
\end{minipage}
\begin{minipage}{0.2\linewidth}
\centerline{\includegraphics[scale=0.2]{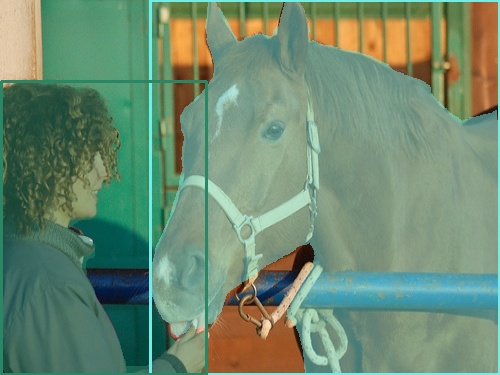}}
\end{minipage}
\begin{minipage}{0.2\linewidth}
\centerline{\includegraphics[scale=0.2]{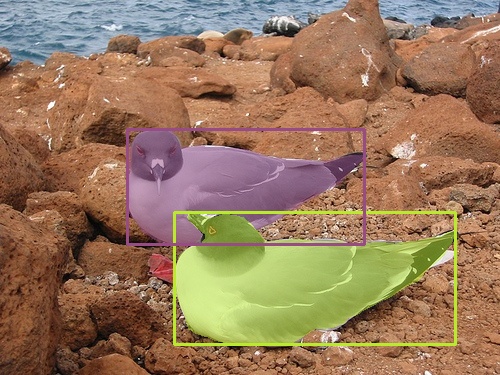}}
\end{minipage}
\begin{minipage}{0.2\linewidth}
\centerline{\includegraphics[scale=0.2]{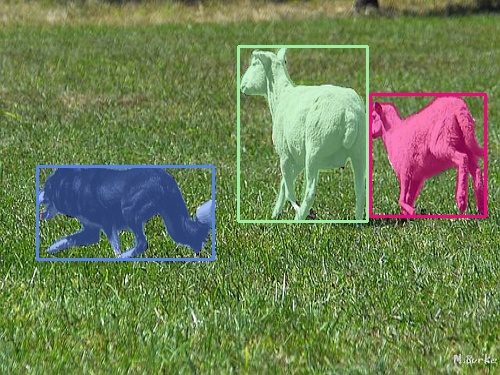}}
\end{minipage}
\\
\vspace{0.1cm}
\begin{minipage}{0.03\linewidth}
\centerline{(c)}
\end{minipage}
\begin{minipage}{0.2\linewidth}
\centerline{\includegraphics[scale=0.2]{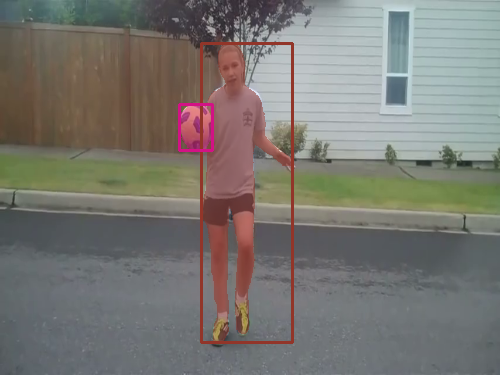}}
\end{minipage}
\begin{minipage}{0.2\linewidth}
\centerline{\includegraphics[scale=0.2]{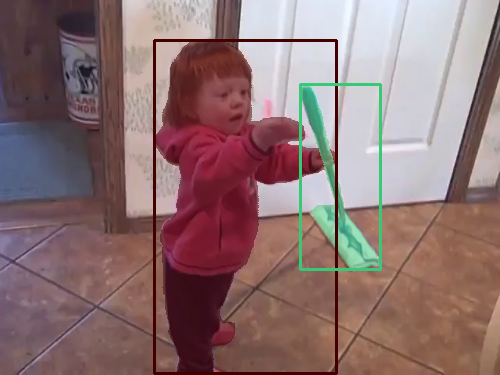}}
\end{minipage}
\begin{minipage}{0.2\linewidth}
\centerline{\includegraphics[scale=0.2]{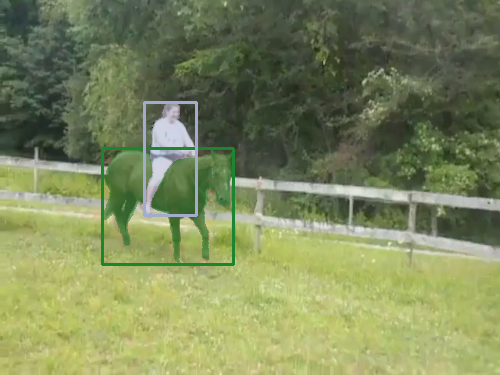}}
\end{minipage}
\begin{minipage}{0.2\linewidth}
\centerline{\includegraphics[scale=0.2]{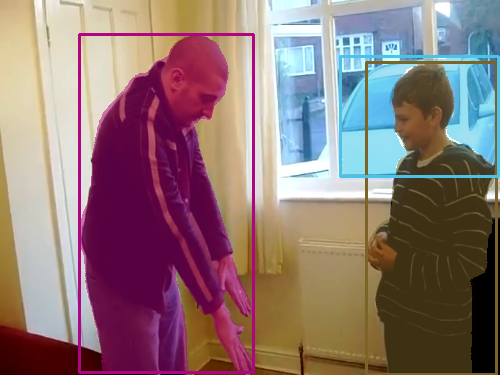}}
\end{minipage}
\\
\vspace{0.1cm}
\begin{minipage}{0.03\linewidth}
\centerline{(d)}
\end{minipage}
\begin{minipage}{0.2\linewidth}
\centerline{\includegraphics[scale=0.2]{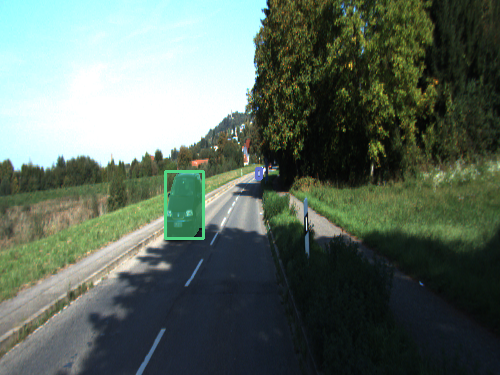}}
\end{minipage}
\begin{minipage}{0.2\linewidth}
\centerline{\includegraphics[scale=0.2]{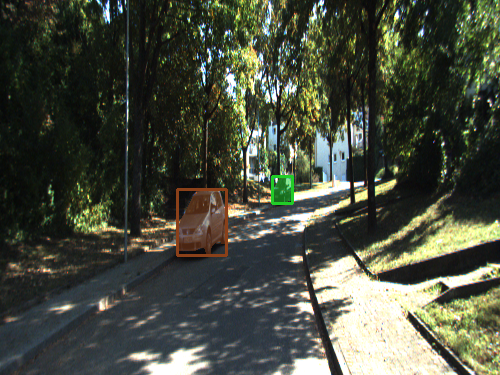}}
\end{minipage}
\begin{minipage}{0.2\linewidth}
\centerline{\includegraphics[scale=0.2]{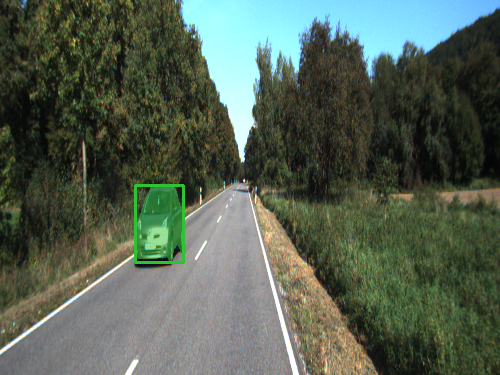}}
\end{minipage}
\begin{minipage}{0.2\linewidth}
\centerline{\includegraphics[scale=0.2]{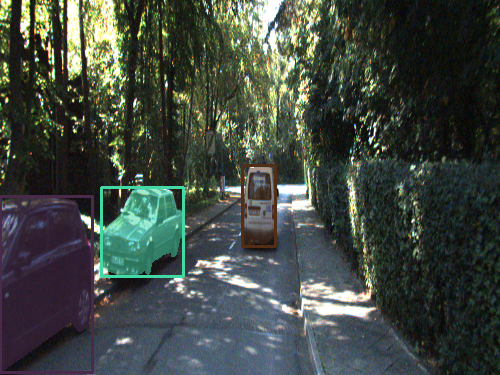}}
\end{minipage}
\\
\vspace{0.1cm}
\begin{minipage}{0.03\linewidth}
\centerline{(e)}
\end{minipage}
\begin{minipage}{0.2\linewidth}
\centerline{\includegraphics[scale=0.2]{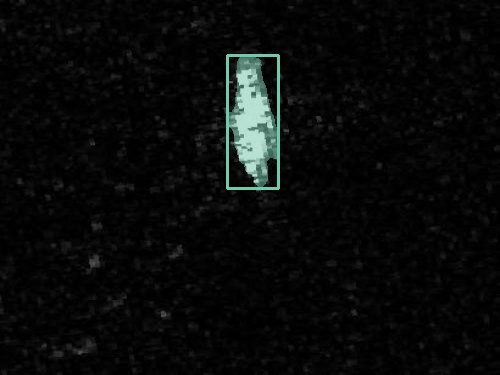}}
\end{minipage}
\begin{minipage}{0.2\linewidth}
\centerline{\includegraphics[scale=0.2]{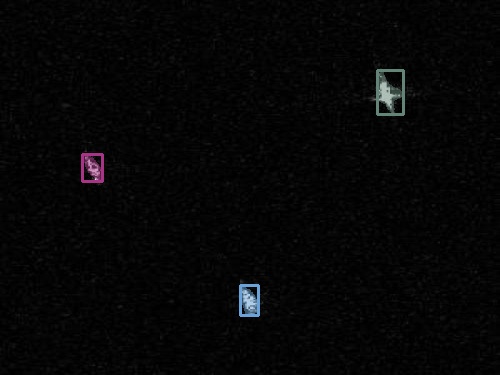}}
\end{minipage}
\begin{minipage}{0.2\linewidth}
\centerline{\includegraphics[scale=0.2]{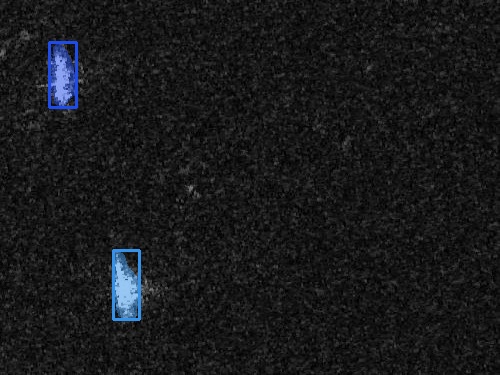}}
\end{minipage}
\begin{minipage}{0.2\linewidth}
\centerline{\includegraphics[scale=0.2]{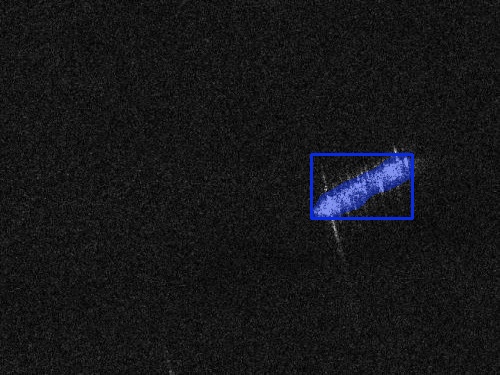}}
\end{minipage}

\caption{More qualitative results. (a) COCO20K \cite{simeoni2021localizing}; (b) Pascal VOC~\citep{everingham2010pascal}; (c) UVO \cite{wang2021unidentified}; (d) KITTI \cite{geiger2012we}; (e) SSDD \cite{zhang2021sar}.}
\label{fig:more}
\end{center}
\end{figure*}

\fref{fig:result_cocoval} shows qualitative results on COCO val2017 \cite{lin2014microsoft}. Each row, from top to bottom, shows the results of CutLER \cite{wang2023cut}, unMORE \cite{yafeiunmore} and our model. The results show that our model produces high-quality object segments with precise boundaries by employing superpixel-guided mask loss. Moreover, our model significantly reduces noise and segments objects with high semantic information by exploiting global context through superpixels. \fref{fig:more} shows more qualitative results of our model for COCO 20K \cite{simeoni2021localizing}, Pascal VOC \cite{everingham2010pascal}, UVO \cite{wang2021unidentified}, KITTI \cite{geiger2012we}, and SSDD \cite{zhang2021sar}.

\subsection{Analysis}

We conduct ablation studies to evaluate the impact of individual components in our framework.

\begin{table*}[!t]
\footnotesize
\begin{center}
\caption{Ablation study on components of the proposed framework using COCO val2017~\cite{lin2014microsoft}.}
\label{tab:ablation}
\begin{tabu}{>{\centering}m{0.07\textwidth}>{\centering}m{0.13\textwidth}>{\centering}m{0.035\textwidth}>{\centering}m{0.035\textwidth}>{\centering}m{0.035\textwidth} | >{\centering}m{0.05\textwidth}>{\centering}m{0.05\textwidth}>{\centering}m{0.05\textwidth}>{\centering}m{0.05\textwidth}>{\centering}m{0.05\textwidth}>{\centering}m{0.05\textwidth} } 
 \hline
 Baseline & Mask Filter & $\calL_{hard}$ & $\calL_{soft}$ &  $\calL_{ad}$ &$\text{AP}^{\text{box}}_{\text{50}}$ & $\text{AP}^{\text{box}}_{\text{75}}$ & $\text{AP}^{\text{box}}$ & $\text{AP}^{\text{mask}}_{\text{50}}$ & $\text{AP}^{\text{mask}}_{\text{75}}$ & $\text{AP}^{\text{mask}}$  \\ 
 \hline \hline
 $\checkmark$ & & & & & 21.9 & 9.6 & 10.6 & 15.4& 6.7 &7.8 \\ 
 $\checkmark$ & $\checkmark$ & & & & 23.8 & 10.5 & 11.7 & 17.5& 7.8& 8.6 \\
 $\checkmark$ & $\checkmark$ & $\checkmark$ & & & 27.5 & 12.8 & 14.1 & 21.3& 10.1 &10.7 \\
 $\checkmark$ & $\checkmark$ & $\checkmark$ & $\checkmark$ & & 31.7 & 15.4 & 16.6 & 25.5&12.4 &13.1 \\ 
 $\checkmark$ & $\checkmark$ & $\checkmark$ & $\checkmark$ & $\checkmark$ & \textbf{34.5} & \textbf{16.7} & \textbf{18.4} & \textbf{28.6}& \textbf{13.8}& \textbf{14.6} \\ 
 \hline
\end{tabu}
\end{center}
\end{table*}

\tref{tab:ablation} shows an ablation study on the components of the proposed framework using COCO val2017 \cite{lin2014microsoft}. The baseline model represents the framework that uses coarse masks generated by RAMA \cite{abbas2022rama} without the Mask Filter for training the segmentation network by the standard mask loss. The predicted masks are then used for self-training the segmentation network to improve its performance. By applying the Mask Filter to obtain high-quality coarse masks, the model achieves better results with increases of 1.1$\%$ in $\text{AP}^{\text{box}}$ and 0.8$\%$ in $\text{AP}^{\text{mask}}$. When combining superpixels and coarse masks to compute the hard loss ($\calL_{hard}$), $\text{AP}^{\text{box}}$ and $\text{AP}^{\text{mask}}$ increase by 2.4$\%$ and 2.1$\%$, respectively. With soft loss ($\calL_{soft}$) to capture global context cues through superpixels, the results are further increased by 2.5$\%$ in $\text{AP}^{\text{box}}$ and 2.4$\%$ in $\text{AP}^{\text{mask}}$. Lastly, applying adaptive loss ($\calL_{ad}$) to efficiently refine the self-training process yields additional improvements, with $\text{AP}^{\text{box}}$ and $\text{AP}^{\text{mask}}$ increasing by 1.8$\%$ and 1.5$\%$, respectively.

\begin{figure*}[!t] 
\begin{center}
\begin{minipage}{0.03\linewidth}
\centerline{(a)}
\end{minipage}
\begin{minipage}{0.2\linewidth}
\centerline{\includegraphics[scale=0.2]{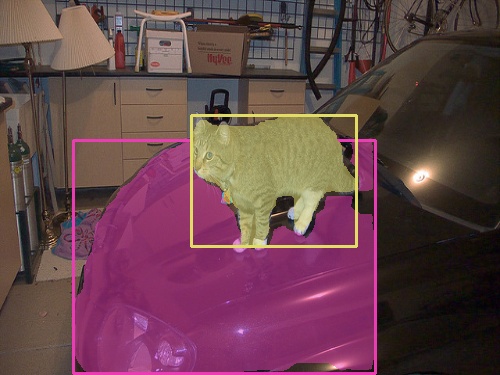}}
\end{minipage}
\begin{minipage}{0.2\linewidth}
\centerline{\includegraphics[scale=0.2]{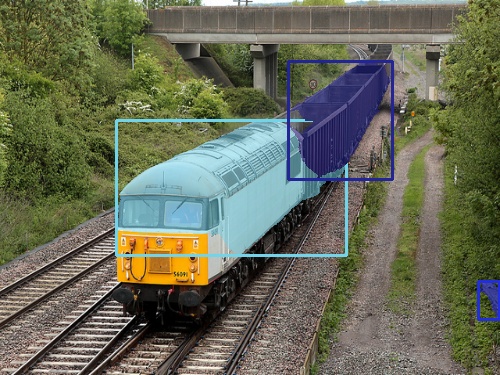}}
\end{minipage}
\begin{minipage}{0.2\linewidth}
\centerline{\includegraphics[scale=0.2]{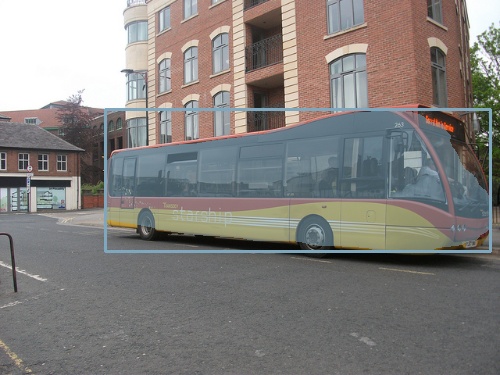}}
\end{minipage}
\begin{minipage}{0.2\linewidth}
\centerline{\includegraphics[scale=0.2]{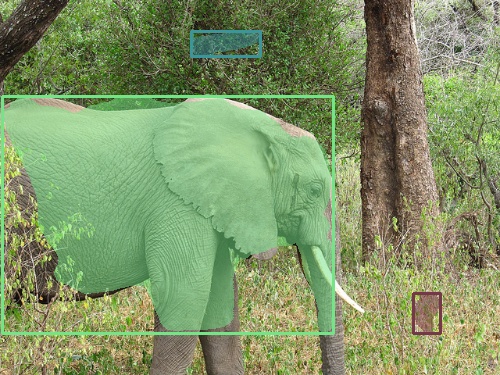}}
\end{minipage}
\\
\vspace{0.1cm}
\begin{minipage}{0.03\linewidth}
\centerline{(b)}
\end{minipage}
\begin{minipage}{0.2\linewidth}
\centerline{\includegraphics[scale=0.2]{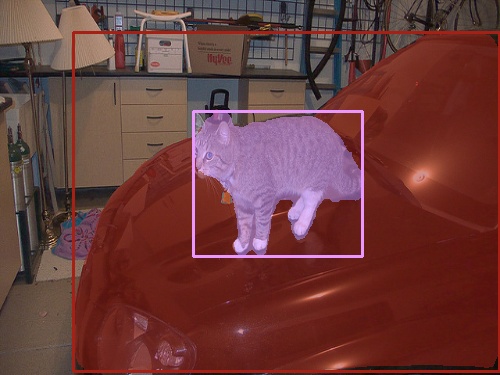}}
\end{minipage}
\begin{minipage}{0.2\linewidth}
\centerline{\includegraphics[scale=0.2]{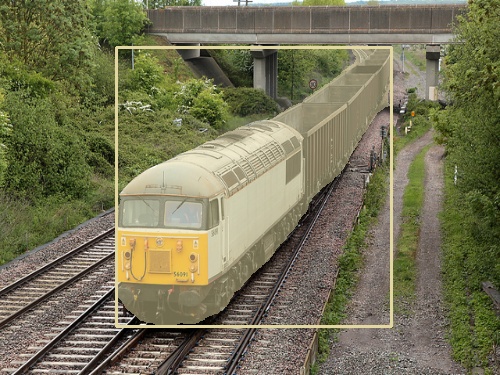}}
\end{minipage}
\begin{minipage}{0.2\linewidth}
\centerline{\includegraphics[scale=0.2]{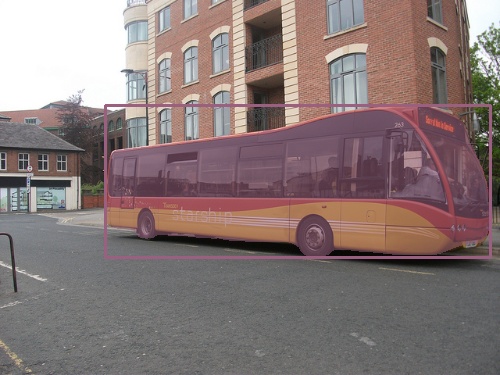}}
\end{minipage}
\begin{minipage}{0.2\linewidth}
\centerline{\includegraphics[scale=0.2]{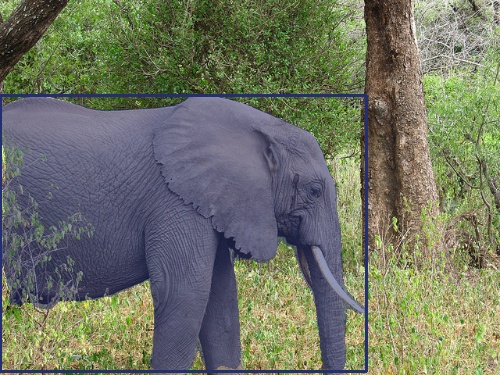}}
\end{minipage}

\caption{Qualitative results on COCO val2017 \cite{lin2014microsoft}. (a) Without $\calL_{sgm}$. (b) With $\calL_{sgm}$.}
\label{fig:ablation}
\end{center}
\end{figure*}

\begin{table*}[!t]
\centering 
\footnotesize
\caption{Comparison of the proposed framework without and with superpixels using COCO val2017~\cite{lin2014microsoft}.}
\label{tab:withsp}
\begin{tabular}{>{\raggedright}m{0.2\textwidth}|>{\centering}m{0.067\textwidth}>{\centering}m{0.067\textwidth}>{\centering\arraybackslash}m{0.067\textwidth}>{\centering\arraybackslash}m{0.067\textwidth}>{\centering\arraybackslash}m{0.067\textwidth}>{\centering\arraybackslash}m{0.067\textwidth}} 
 \hline
 Methods &$\text{AP}^{\text{box}}_{\text{50}}$ & $\text{AP}^{\text{box}}_{\text{75}}$ & $\text{AP}^{\text{box}}$ & $\text{AP}^{\text{mask}}_{\text{50}}$ & $\text{AP}^{\text{mask}}_{\text{75}}$ & $\text{AP}^{\text{mask}}$  \\ 
 \hline \hline
  w/o superpixels & 28.2 & 12.3 & 13.8 & 22.7& 9.5 &10.9 \\ 
 w/ superpixels & \textbf{34.5} & \textbf{16.7} & \textbf{18.4} & \textbf{28.6}& \textbf{13.8}& \textbf{14.6} \\ 
 \hline
\end{tabular}
\end{table*}
\begin{table*}[!t]
\footnotesize
\begin{center}
\caption{Comparison of coarse mask generation methods on COCO val2017~\cite{lin2014microsoft}.}
\label{tab:rama}
\begin{tabular}{>{\raggedright}m{0.2\textwidth}|>{\centering}m{0.067\textwidth}>{\centering}m{0.067\textwidth}>{\centering\arraybackslash}m{0.067\textwidth}>{\centering\arraybackslash}m{0.067\textwidth}>{\centering\arraybackslash}m{0.067\textwidth}>{\centering\arraybackslash}m{0.067\textwidth}} 
 \hline
Methods & $\text{AP}^{\text{box}}_{\text{50}}$ & $\text{AP}^{\text{box}}_{\text{75}}$ & $\text{AP}^{\text{box}}$ & $\text{AP}^{\text{mask}}_{\text{50}}$ & $\text{AP}^{\text{mask}}_{\text{75}}$ & $\text{AP}^{\text{mask}}$\\ 
\hline\hline

FreeMask~\citep{wang2022freesolo} &25.6  & 12.5 & 15.4& 20.4 & 10.0 & 11.2 \\ 
MaskCut~\citep{wang2023cut} & 30.3 & 15.4 & 17.1& 26.1 & 12.1 & 13.5\\ 
RAMA \cite{abbas2022rama} (Ours) & \textbf{34.5} & \textbf{16.7} & \textbf{18.4} & \textbf{28.6}& \textbf{13.8}& \textbf{14.6}\\ 
\hline
\end{tabular}
\end{center}
\end{table*}

\begin{table*}[!t]
\footnotesize
\begin{center}
\caption{Comparison of superpixel segmentation methods on COCO val2017~\cite{lin2014microsoft}.}
\label{tab:superpixel}
\begin{tabular}{>{\raggedright}m{0.2\textwidth}|>{\centering}m{0.067\textwidth}>{\centering}m{0.067\textwidth}>{\centering\arraybackslash}m{0.067\textwidth}>{\centering\arraybackslash}m{0.067\textwidth}>{\centering\arraybackslash}m{0.067\textwidth}>{\centering\arraybackslash}m{0.067\textwidth}} 
 \hline
Methods & $\text{AP}^{\text{box}}_{\text{50}}$ & $\text{AP}^{\text{box}}_{\text{75}}$ & $\text{AP}^{\text{box}}$ & $\text{AP}^{\text{mask}}_{\text{50}}$ & $\text{AP}^{\text{mask}}_{\text{75}}$ & $\text{AP}^{\text{mask}}$\\ 
\hline\hline
SNIC \cite{achanta2017superpixels} & 32.9 & 15.3 & 17.1& 26.2 & 12.3 &  13.5 \\ 
GMMSP~\citep{ban2018superpixel} & 33.2 & 15.9 &17.5 &26.7 & 12.8& 13.9  \\ 
LRW~\citep{kang2020dynamic}& 33.7 & 16.2&17.9 & 27.9 & 13.5 & 14.2 \\ 
MCG \cite{arbelaez2014multiscale} (Ours) & \textbf{34.5} & \textbf{16.7} & \textbf{18.4} & \textbf{28.6}& \textbf{13.8}& \textbf{14.6} \\ 
\hline
\end{tabular}
\end{center}
\end{table*}

\fref{fig:ablation} shows qualitative results of our models without and with superpixel-guided mask loss ($\calL_{sgm}$). Without the superpixel-guided mask loss, the segmentation network is trained only with coarse masks using the standard mask loss. The result shows that $\calL_{sgm}$ improve the performance significantly by leveraging advantages of both superpixels and coarse masks.   

\begin{table*}[!t]
\footnotesize
\begin{center}
\caption{Comparison of self-training methods on COCO val2017~\cite{lin2014microsoft}.}
\label{tab:self}
\begin{tabular}{>{\raggedright}m{0.2\textwidth}|>{\centering}m{0.067\textwidth}>{\centering}m{0.067\textwidth}>{\centering\arraybackslash}m{0.067\textwidth}>{\centering\arraybackslash}m{0.067\textwidth}>{\centering\arraybackslash}m{0.067\textwidth}>{\centering\arraybackslash}m{0.067\textwidth}} 
 \hline
Methods & $\text{AP}^{\text{box}}_{\text{50}}$ & $\text{AP}^{\text{box}}_{\text{75}}$ & $\text{AP}^{\text{box}}$ & $\text{AP}^{\text{mask}}_{\text{50}}$ & $\text{AP}^{\text{mask}}_{\text{75}}$ & $\text{AP}^{\text{mask}}$\\ 
\hline\hline
ST++~\citep{yang2022st++} & 33.4 & 15.9 & 17.4 & 26.7 & 13.0 & 13.5 \\ 
Multi-Round~\citep{wang2023cut} & 33.8 & 16.0 & 17.7 & 27.4 & 13.3 & 14.0 \\ 
Ours & \textbf{34.5} & \textbf{16.7} & \textbf{18.4} & \textbf{28.6}& \textbf{13.8}& \textbf{14.6} \\ 
\hline
\end{tabular}
\end{center}
\end{table*}

\begin{table*}[t]
\centering
\footnotesize
\caption{Ablation studies for hyperparameters on COCO val2017~\cite{lin2014microsoft}.}
\label{tab:hyper}
\begin{tabular}{cccccc}
\begin{minipage}[t]{0.2\textwidth}
    \centering
    (a) $Q$ for mask filter\\
    \begin{tabular}{c|c}
    \hline
    $Q\%$ & $\text{AP}^{\text{mask}}$ \\
    \hline
    \hline
    40 & 13.7 \\
    50 & 14.1 \\
    60 & \textbf{14.6} \\
    70 & 13.8 \\
    80 & 13.2 \\
    \hline
    \end{tabular}
\end{minipage} &
\begin{minipage}[t]{0.2\textwidth}
    \centering
    (b) $\alpha_1$ for hard loss \\
    \begin{tabular}{c|c}
    \hline
    $\alpha_1$ & $\text{AP}^{\text{mask}}$ \\
    \hline
    \hline
    50 & 13.9 \\
    100 & \textbf{14.6} \\
    150 & 14.3 \\
    200 & 13.7 \\
    250 & 12.9 \\
    \hline
    \end{tabular}
\end{minipage} &
\begin{minipage}[t]{0.2\textwidth}
    \centering
    (c) $\alpha_2$ for soft loss  \\
    \begin{tabular}{c|c}
    \hline
     $\alpha_2$ & $\text{AP}^{\text{mask}}$ \\
    \hline
    \hline
    50 & 13.1 \\
    100 & 13.4 \\
    150 & 14.2 \\
    200 & \textbf{14.6} \\
    250 & 14.0 \\
    \hline
    \end{tabular}
\end{minipage} &
\\
\\
\begin{minipage}[t]{0.2\textwidth}
    \centering
    (d) $e$ for adaptive loss  \\
    \begin{tabular}{c|c}
    \hline
     $e$ & $\text{AP}^{\text{mask}}$ \\
    \hline
    \hline
    2 & 13.6 \\
    3 & \textbf{14.6} \\
    4 & 14.1 \\
    5 & 13.3 \\
    6 & 12.8 \\
    \hline
    \end{tabular}
\end{minipage} &
\begin{minipage}[t]{0.2\textwidth}
    \centering
     (e) $\epsilon$ for adaptive loss \\
    \begin{tabular}{c|c}
    \hline
     $\epsilon$ & $\text{AP}^{\text{mask}}$ \\
    \hline
    \hline
    0.4 & 13.2 \\
    0.5 & 14.1 \\
    0.6 & \textbf{14.6} \\
    0.7 & 14.3 \\
    0.8 & 13.5 \\
    \hline
    \end{tabular}
\end{minipage} &
\begin{minipage}[t]{0.2\textwidth}
    \centering
    (f) $\hat{d}$ for adaptive loss \\
    \begin{tabular}{c|c}
    \hline
     $\hat{d}$ & $\text{AP}^{\text{mask}}$ \\
    \hline
    \hline
    1 & 13.5 \\
    2 & 13.9 \\
    3 & \textbf{14.6} \\
    4 & 14.2 \\
    5 & 13.5 \\
    \hline
    \end{tabular}
\end{minipage}
\end{tabular}
\end{table*}

\begin{table*}[!t]
\footnotesize
\begin{center}
\caption{Impact of dataset characteristics used to train DINO and our segmentation network (SN). Evaluation is conducted on COCO val2017~\cite{lin2014microsoft}.}
\label{tab:impact}
\begin{tabular}{>{\raggedright}m{0.1\textwidth}>{\raggedright}m{0.1\textwidth}|>{\centering}m{0.067\textwidth}>{\centering}m{0.067\textwidth}>{\centering\arraybackslash}m{0.067\textwidth}>{\centering\arraybackslash}m{0.067\textwidth}>{\centering\arraybackslash}m{0.067\textwidth}>{\centering\arraybackslash}m{0.067\textwidth}} 
 \hline
 DINO \cite{caron2021emerging}& SN &$\text{AP}^{\text{box}}_{\text{50}}$ & $\text{AP}^{\text{box}}_{\text{75}}$ & $\text{AP}^{\text{box}}$ & $\text{AP}^{\text{mask}}_{\text{50}}$ & $\text{AP}^{\text{mask}}_{\text{75}}$ & $\text{AP}^{\text{mask}}$  \\ 
 \hline \hline
YFCC1M &IN1K  & 25.1 & 9.7 & 12.9 & 19.8 &7.6&9.7 \\ 
IN1K&YFCC1M & 25.7 & 10.4 & 13.3 &21.1&8.3& 10.2 \\ 
YFCC1M &YFCC1M  & 30.6 & 13.9 & 16.7 & 26.1 & 11.5 & 12.9 \\ 
 IN1K&IN1K  & \textbf{31.3} & \textbf{14.6} & \textbf{17.5} & \textbf{26.8}& \textbf{12.1}& \textbf{13.4} \\ 
 \hline
\end{tabular}
\end{center}
\end{table*}

\tref{tab:withsp} shows the effectiveness of superpixels in our framework. In the proposed framework without superpixels, we compute $\mathcal{L}_{hard}$ as a standard mask loss using only coarse masks and calculate $\mathcal{L}_{soft}$ at the pixel level by constructing the graph $\mathcal{G}$ with pixels as nodes. The results show that applying superpixels to train the segmentation network increases $\text{AP}^{\text{box}}$ and $\text{AP}^{\text{mask}}$ by 4.6$\%$ and 3.7$\%$, respectively. This indicates that superpixels significantly improve the performance of our model.

\tref{tab:rama} presents the results of our method using various coarse mask generation models, including FreeMask in FreeSOLO \cite{wang2022freesolo} and MaskCut in CutLER \cite{wang2023cut}. The results demonstrate that our model consistently outperforms previous state-of-the-art methods, even when using their coarse mask generation techniques. This proves the effectiveness of our contributions.

\begin{table*}[!t]
\footnotesize
\begin{center}
\caption{Comparison of the proposed framework with other instance segmentation models using COCO val2017~\cite{lin2014microsoft}.}
\label{tab:instancemodel}
\begin{tabular}{>{\raggedright}m{0.2\textwidth}|>{\centering}m{0.067\textwidth}>{\centering}m{0.067\textwidth}>{\centering\arraybackslash}m{0.067\textwidth}>{\centering\arraybackslash}m{0.067\textwidth}>{\centering\arraybackslash}m{0.067\textwidth}>{\centering\arraybackslash}m{0.067\textwidth}} 
 \hline
 Methods &$\text{AP}^{\text{box}}_{\text{50}}$ & $\text{AP}^{\text{box}}_{\text{75}}$ & $\text{AP}^{\text{box}}$ & $\text{AP}^{\text{mask}}_{\text{50}}$ & $\text{AP}^{\text{mask}}_{\text{75}}$ & $\text{AP}^{\text{mask}}$  \\ 
 \hline \hline
  Mask2Former \cite{cheng2022masked} & 24.4 & 11.2 & 12.5 & 18.3& 7.7 &9.3 \\ 
Mask DINO \cite{li2023mask} & 28.2 & 12.4 & 14.3 & 22.5& 9.3 &10.9 \\ 
 Ours & \textbf{34.5} & \textbf{16.7} & \textbf{18.4} & \textbf{28.6}& \textbf{13.8}& \textbf{14.6} \\ 
 \hline
\end{tabular}
\end{center}
\end{table*}

\tref{tab:superpixel} shows the results of our method using various superpixel segmentation models, including SNIC \cite{achanta2017superpixels}, GMMSP~\citep{ban2018superpixel}, LRW~\citep{kang2020dynamic}, and MCG \cite{arbelaez2014multiscale}. These methods are efficient and easy to implement. The results show that our model brings good performances with other superpixel segmentation algorithms.

\begin{table*}[t!]
\footnotesize
\centering
\caption{Unsupervised universal segmentation COCO val2017~\cite{lin2014microsoft}.}
\label{tab:universal}
\begin{tabular}{>{\raggedright}m{0.12\textwidth}|>{\centering}m{0.066\textwidth}>{\centering}m{0.066\textwidth}|>{\centering}m{0.066\textwidth}>{\centering}m{0.066\textwidth}|>{\centering\arraybackslash}m{0.066\textwidth}>{\centering}m{0.05\textwidth}|>{\centering\arraybackslash}m{0.05\textwidth}>{\centering}m{0.05\textwidth}>{\centering\arraybackslash}m{0.05\textwidth}>{\centering}m{0.03\textwidth}>{\centering\arraybackslash}m{0.03\textwidth}>{\centering}m{0.03\textwidth}}  
\hline
Task & \multicolumn{2}{c|}{Agn Instance Seg.} & \multicolumn{2}{c|}{Instance Seg.}& \multicolumn{2}{c|}{Semantic Seg.}& \multicolumn{3}{c}{Panoptic Seg.} \\

 Metric& $\text{AP}^{\text{box}}_{\text{50}}$ & $\text{AP}^{\text{box}}$ & $\text{AP}^{\text{box}}_{\text{50}}$ & $\text{AR}^{\text{box}}_{\text{100}}$ & PixelAcc & mIoU & PQ & SQ & RQ \\
\hline\hline
U2Seg \cite{niu2024unsupervised}& 22.8 & 13.0 & 11.8 & 21.5& 63.9 & 30.2 & 16.1 & 71.1 & 19.9   \\ 
Ours& \textbf{34.9} & \textbf{18.1} & \textbf{18.3} & \textbf{40.6}& \textbf{72.3} & \textbf{34.5}& \textbf{20.6} & \textbf{79.5} & \textbf{29.4}   \\

 \hline
\end{tabular}

\end{table*}

\begin{figure*}[t!] 
\begin{center}

\begin{minipage}{0.03\linewidth}
\centerline{(a)}
\end{minipage}
\begin{minipage}{0.2\linewidth}
\centerline{\includegraphics[scale=0.2]{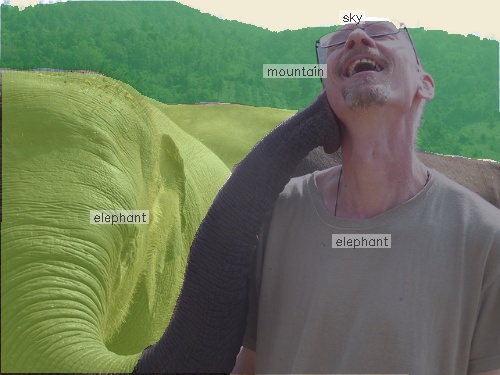}}
\end{minipage}
\begin{minipage}{0.2\linewidth}
\centerline{\includegraphics[scale=0.2]{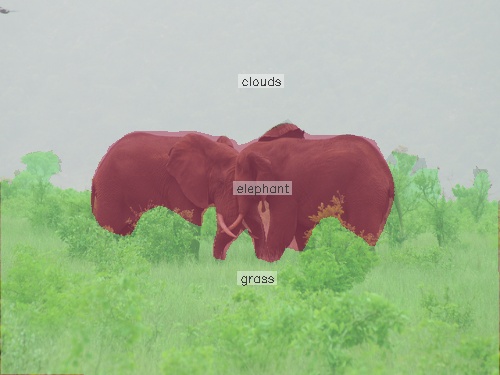}}
\end{minipage}
\begin{minipage}{0.2\linewidth}
\centerline{\includegraphics[scale=0.2]{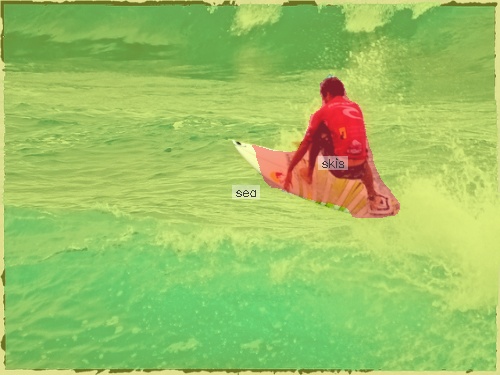}}
\end{minipage}
\begin{minipage}{0.2\linewidth}
\centerline{\includegraphics[scale=0.2]{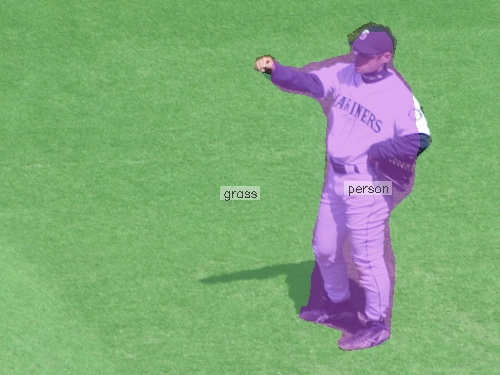}}
\end{minipage}
\\
\vspace{0.1cm}
\begin{minipage}{0.03\linewidth}
\centerline{(b)}
\end{minipage}
\begin{minipage}{0.2\linewidth}
\centerline{\includegraphics[scale=0.2]{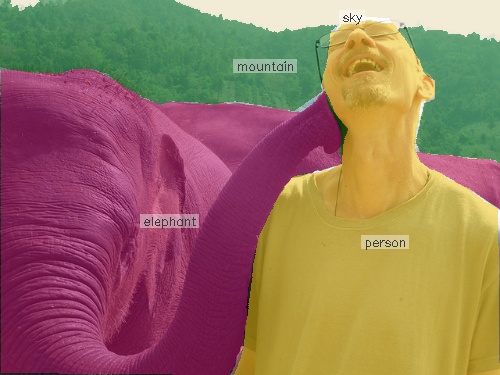}}
\end{minipage}
\begin{minipage}{0.2\linewidth}
\centerline{\includegraphics[scale=0.2]{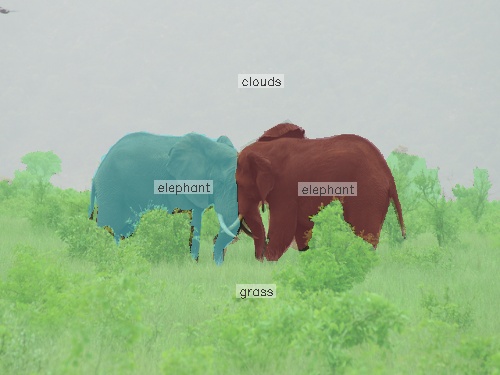}}
\end{minipage}
\begin{minipage}{0.2\linewidth}
\centerline{\includegraphics[scale=0.2]{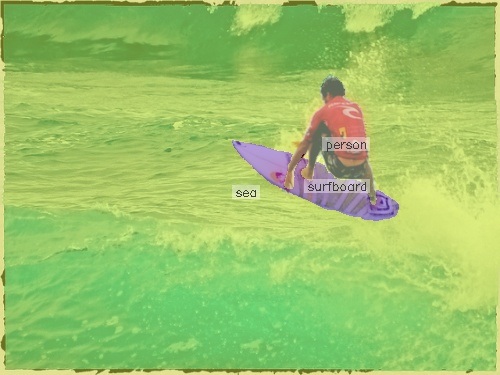}}
\end{minipage}
\begin{minipage}{0.2\linewidth}
\centerline{\includegraphics[scale=0.2]{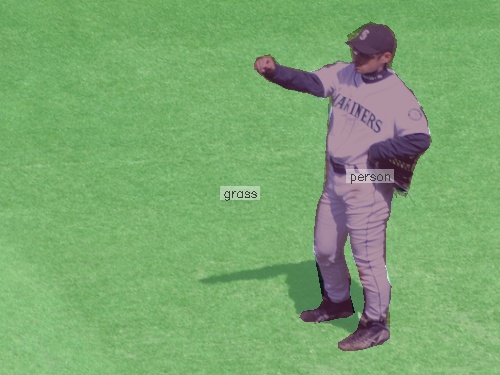}}
\end{minipage}

\caption{Qualitative results for unsupervised universal image segmentation on COCO val2017 \cite{lin2014microsoft}. (a) U2Seg \cite{niu2024unsupervised}. (b) Ours.}
\label{fig:papno}
\end{center}
\end{figure*}

\tref{tab:self} compares our adaptive loss ($\calL_{ad}$) with ST++~\citep{yang2022st++} and Multi-Round Self-Training \cite{wang2023cut}. Here, ST++~\citep{yang2022st++} also utilizes the holistic stability of predicted masks to evaluate their reliability. However, they use a fixed confidence threshold to filter out 
low-confidence masks. For Multi-Round Self-Training in CutLER \cite{wang2023cut}, multiple rounds of self-training are performed on the predictions of the model to produce finer segmentation masks. The results demonstrate that our adaptive loss ($\calL_{ad}$) obtains the best performance by adaptively weighting the loss of masks based on their reliability, allowing the model to fully leverage advantages of reliable masks and mitigate negative impact of unreliable masks. Furthermore, with adaptive loss ($\calL_{ad}$), the model requires only one round of self-training, which significantly reduces training time.

In \tref{tab:hyper}, we present an analysis on the hyperparameters. The results show that the proposed method consistently surpasses prior approaches, even when less optimal hyperparameter values are employed.

To assess the impact of the dataset characteristics, \tref{tab:impact} shows the performance of our method using an object-centric dataset, ImageNet \cite{deng2009imagenet}, and a non-object-centric dataset, YFCC \cite{thomee2016yfcc100m}, to pre-train DINO \cite{caron2021emerging} and to train our segmentation network. Following CutLER~\citep{wang2023cut}, we control the number of images in both datasets to ensure a fair comparison. The performance is evaluated on COCO val2017~\cite{lin2014microsoft}. The table demonstrates that our model maintains robust performance when the same dataset is used to train both DINO \cite{caron2021emerging} and the segmentation network, highlighting its generalization capability. Nevertheless, training on different datasets results in degraded performance, emphasizing the importance of consistent image characteristics when training both DINO \cite{caron2021emerging} and the segmentation network.

\tref{tab:instancemodel} compares our model with other instance segmentation models, including Mask2Former \cite{cheng2022masked} and Mask DINO \cite{li2023mask}. These models are trained using SwinL \cite{liu2021swin} as the backbone and the generated coarse masks as supervision. The results show that our model outperforms Mask2Former \cite{cheng2022masked} by 5.9$\%$ in $\text{AP}^{\text{box}}$ and 5.3$\%$ in $\text{AP}^{\text{mask}}$. Compared to Mask DINO \cite{li2023mask}, our model achieves improvements of 4.1$\%$ in $\text{AP}^{\text{box}}$ and 3.7$\%$ in $\text{AP}^{\text{mask}}$. These results demonstrate the effectiveness of our model when trained with coarse masks.

\subsection{Extension to Unsupervised Universal Segmentation}

In \tref{tab:universal}, we show the additional values of the proposed method by applying it to a new setting called unsupervised universal image segmentation. In the current work U2Seg \cite{niu2024unsupervised}, CutLER \cite{wang2023cut} is combined with unsupervised semantic segmentation model STEGO \cite{hamilton2022unsupervised} to generate pseudo labels for training the universal image segmentation model. Then, the optimized model can perform various tasks, including instance, semantic and panoptic segmentation. To demonstrate our effectiveness in unsupervised universal segmentation, we simply replace CutLER \cite{wang2023cut} with our model for evaluation. All experimental settings are based on U2Seg \cite{niu2024unsupervised}. Firstly, in agnostic instance segmentation, our model outperforms U2Seg by 12.1$\%$ in $\text{AP}^{\text{box}}_{50}$ and by 5.1$\%$ in $\text{AP}^{\text{box}}$. In unsupervised instance segmentation, our model surpasses U2Seg by 6.5$\%$ in $\text{AP}^{\text{box}}_{50}$ and 19.1$\%$ in $\text{AP}^{\text{box}}_{100}$. For unsupervised semantic segmentation, our model achieves 8.4$\%$ higher PixelAcc and 4.3$\%$ higher mIoU compared to U2Seg. In unsupervised panoptic segmentation, our model leads U2Seg by 4.5$\%$ in PQ, 8.4$\%$ in SQ, and 9.5$\%$ in RQ.

\fref{fig:papno} shows qualitative results for unsupervised universal segmentation on COCO val2017 \cite{lin2014microsoft}. The results clearly demonstrate that our model produces higher-quality object masks, whereas U2Seg \cite{niu2024unsupervised} with CutLER \cite{wang2023cut} yields lower-quality object masks, leading to incorrect category assignments.

\section{CONCLUSION}
We presented a novel framework for unsupervised instance segmentation. To train the segmentation network without human annotations, we leveraged easily accessible forms of supervision, including superpixels, coarse masks, and image colors, to improve performance. We proposed a novel superpixel-guided mask loss that leverages color pairwise affinities to estimate the probability of a superpixel being foreground. This loss comprises two components. The first component, called hard loss, is computed by converting coarse masks into hard labels for superpixels. The second component, called soft loss, is calculated by capturing global pairwise affinities of superpixel corlors to produce soft labels for superpixels. Lastly, we proposed the self-training process with the new adaptive loss that utilizes the holistic stability of the predicted masks to efficiently improve their quality. The experimental results indicate that the proposed method outperforms previous state-of-the-art methods on public datasets.

The limitation of the proposed model is that it only focuses on segmenting objects masks while ignoring their categories. In the future, we plan to design a framework that can effectively provide precise masks along with categories for instances, without the need for human annotations during training.  



\section*{Acknowledgments}
The author would like to express sincere thanks to the Editor-in-Chief Professor Zoran Duric, Professor Petia Radeva, the Editors and anonymous reviewers for their valuable comments and suggestions, which greatly improved this paper.

\bibliographystyle{elsarticle-harv}
\bibliography{2_2_mybibfile1}

\end{document}